\documentclass[journal]{IEEEtran}
\ifCLASSINFOpdf
\else
\fi
\usepackage{graphicx}
\usepackage{amsmath}
\usepackage{graphics}
\usepackage{amsfonts}
\usepackage{amssymb}%
\usepackage{amsthm}
\usepackage{hyperref}
\usepackage{epsfig}
\usepackage{subfigure}
\usepackage{color}
\usepackage{listings}
\usepackage{booktabs}
\usepackage{multirow}
\usepackage{algorithm}

\usepackage{algorithmic}
\usepackage{amsfonts}
\usepackage{multicol}
\usepackage{epstopdf}
\usepackage[justification=centering]{caption}

\usepackage{cleveref}

\newtheorem{theorem}{Theorem}
\newtheorem{lemma}{Lemma}

\newtheorem{proposition}{Proposition}

\newtheorem{definition}{Definition}

\newcommand{\tp}{{\tilde{p}}}
\newcommand{\tu}{{\tilde{u}}}
\newcommand{\tv}{{\tilde{v}}}
\newcommand{\tw}{{\tilde{w}}}

\begin{document}
\title{Fast Polynomial Kernel Classification for Massive Data}

\author{Jinshan~Zeng, Minrun Wu, Shao-Bo Lin, Ding-Xuan Zhou % <-this % stops a space
\thanks{J. Zeng and M. Wu are with the School of Computer and Information Engineering, Jiangxi Normal University, Nanchang, China (email: jinshanzeng@jxnu.edu.cn, wuminrun36@gmail.com).}% <-this % stops a space
\thanks{S.B. Lin is with the School of Management, Xi'an Jiaotong University, Xi'an, China (email: sblin1983@gmail.com)}% <-this % stops a space
\thanks{D.X. Zhou is with the School of Mathematics and Statistics, University of Sydney, NSW 2006, Australia (email: dingxuan.zhou@sydney.edu.au)}% <-this % stops a space
\thanks{The corresponding author is Shao-Bo Lin}
}
\maketitle

\begin{abstract}
In the era of big data, it is desired to develop efficient machine learning algorithms to tackle massive data challenges  such as storage bottleneck, algorithmic scalability, and interpretability. In this paper, we develop a novel efficient classification algorithm, called fast polynomial kernel classification (FPC), to conquer the scalability and storage challenges. Our main tools are a suitable selected feature mapping based on polynomial kernels and an alternating direction method of multipliers (ADMM) algorithm for a related non-smooth convex optimization problem. Fast learning rates as well as  feasibility verifications including the efficiency of an ADMM solver with convergence guarantees and the selection of center points are established to justify  theoretical behaviors of FPC. Our theoretical assertions are verified by a series of simulations and real data applications. Numerical results demonstrate that FPC significantly reduces the computational burden and storage memory of existing learning schemes such as support vector machines, Nystr\"{o}m and random feature methods, without sacrificing their generalization abilities much.
\end{abstract}


% \fi
\IEEEpeerreviewmaketitle

\section{Introduction}
With the rapid development of data mining, massive data abound around   our lives in
terms of medical records,  high-frequency financial data,  internet data, network data, longitudinal data, image data and so on. For examples,  millions of Internet URLs
are inspected for   detection of pop-up junk messages;    hundreds of millions of financial records are exploited for predicting   financial trends; and billions of
customer activities  are gathered for making marketing decisions.
These     massive data  make the prediction much more precise and bring opportunities to discover  subtle information which cannot be captured  by data
of small size. However, they simultaneously produce a series of scientific
challenges such as   storage bottleneck, algorithmic
scalability, and interpretability  \cite{Zhou2014}.
In  the machine learning community, developing scalable learning algorithms with theoretical verifications to conquer the massive data challenges is a recent focus and has triggered enormous research activities \cite{Chen2014,Zhou2014}.

For large-scale regression tasks, some scalable learning schemes including localized kernel ridge regression \cite{Meister2016},
distributed learning  \cite{Zhang2015}, learning with sub-sampling \cite{Gittens2016},
have been proposed to tackle massive data. All these schemes are rigorously  justified to significantly  reduce the computational burden of  classical learning schemes such as   kernel methods \cite{Taylor2004} and neural networks \cite{Hagan1996} without sacrificing their generalization performances very much. In fact, it has been  proved in  \cite{Meister2016,Lin2017-DRLS,Lin2017-CA,Rudi2015} that these scalable schemes can achieve the optimal learning rates for kernel approaches in the framework of statistical learning theory \cite{Cucker2007}.

From regression to classification, the  least-square fitting schemes are frequently
replaced by the margin theory \cite{Taylor2004}. As a consequence,  loss functions are changed from least-squares to margin-based functions such as the hinge loss for support vector machine (SVM) \cite{Taylor2004}, logistic loss for logistic regression \cite{Menard2002}, and exponential loss for boosting \cite{Freund1997}. As a result, the proposed approaches for regression are no longer efficient
%no more efficient
for massive data classification. Based on the maximal margin principle, there have been several scalable algorithms for classification, including  distributed SVM \cite{Forero2010} ,  localized SVM \cite{Dumpert2018},  sequential minimal optimization (SMO) \cite{Platt1999},
classification with sub-sampling \cite{Chien2010,Rudi2015,Rudi2017}, SVM with approximated feature maps \cite{Magi2009,Veldaldi2010,Veldaldi2012,Kar2012}, and SVM with random features \cite{Rahimi2007,Vempati2010,Pham2013}.
Although  these classification schemes can reduce the computational complexity of SVM and perform well in some special classification tasks, their performance is sensitive to involved parameters and thus, they generally  require delicate parameter-selection strategies, which usually brings huge computations in the training process. Furthermore,  most of them lack  theoretical verifications on the generalization ability,  which hinders  practitioners' spirits to use them in massive data classifications.

The aim of the present paper is to propose a novel scalable learning algorithm with theoretical verifications for massive data classification. Different from the maximal margin principle, our  basic idea is to select a suitable feature space in which   linear  classifiers with different margins perform similarly.  For this purpose,  we use polynomial kernels to build up the feature mapping and
control   the capacity of feature space via tuning the kernel
parameter. Furthermore, since  the margin is not so important in our approach and can be removed, our method then turns to  solving  a non-smooth convex  optimization problem. We adopt  an alternating direction method of multipliers (ADMM) to solve this problem and then propose a novel learning algorithm, named as fast polynomial kernel classification (FPC) to tackle massive data. Two important advantages of FPC are: (a) since the margin constraint (or regularization parameter)
 in SVM is not required in FPC and capacity of feature space is determined by  the kernel parameter, there  are at most two discrete  parameters to be tuned in FPC; (b) For each fixed kernel parameter, the computational complexity of FPC is much smaller than that of SVM. Both advantages show that FPC can significantly reduce the computational burden of SVM and thus succeeds in tackling massive data.

The excellent performance of FPC is verified  by both theoretical
analysis and numerical experiments.
Theoretically,  we present some  feasibility guarantees for FPC including the expressivity of the feature mapping and  efficiency of the ADMM algorithm with convergence guarantees. Furthermore, we rigorously prove  that under the   Tsybakov noise \cite{Tsybakov2004} and geometric  noise \cite{Steinwart2007} assumptions, FPC achieves the existing  optimal learning rates for SVM in the framework of statistical learning theory \cite{Cucker2007,Guo2020,Steinwart2008}. Experimentally, numerical studies including  toy simulations, UCI standard data experiments, massive data trials and three high-dimensional image classification experiments are conducted to illustrate the outperformance of FPC. Our numerical results show that FPC is more efficient than some state-of-the-art methods such as some existing Nystr\"{o}m and random feature methods \cite{Rudi2017,Rahimi2007,Pham2013}, in the sense that it can significantly reduce the computational burden and storage requirements without degrading the generalization capability much.

The rest of this paper is organized as follows.
In Section \ref{sc:feature-mapping}, we present the motivation of our study.
Section \ref{sc:feasibility} proposes the FPC algorithm as well as some  feasibility verifications.
Section \ref{sc:theoretical-behavior} derives the generalization error estimates for FPC in the framework of learning theory.
Section \ref{sc:simulations} provides a series of toy simulations to verify the feasibility of FPC.
Section \ref{sc:realdata} conducts  a series of UCI data sets and real applications to show the effectiveness of FPC in massive data classification.
We conclude this paper in Section \ref{sc:conclusion}.

\section{Motivation and Roadmap}
\label{sc:feature-mapping}

In this section, we aim at presenting  motivations of our study and providing  a roadmap  to conquer the massive data challenge for binary classification.

\subsection{Motivations}
\label{sc:motivations}
The maximal margin principle is an important tool to  design learning algorithms for binary classification. A margin-based algorithm   explicitly utilizes the margin of each data point to produce an efficient classifier, where  the  margin of a single data point is defined to be the distance from the data point to a decision boundary. In this way, both SVM \cite{Taylor2004} and boosting \cite{Rosset2004} can be regarded as margin-based algorithms.

\begin{figure}[!t]
\begin{minipage}[b]{0.99\linewidth}
\centering
\includegraphics*[scale=0.25]{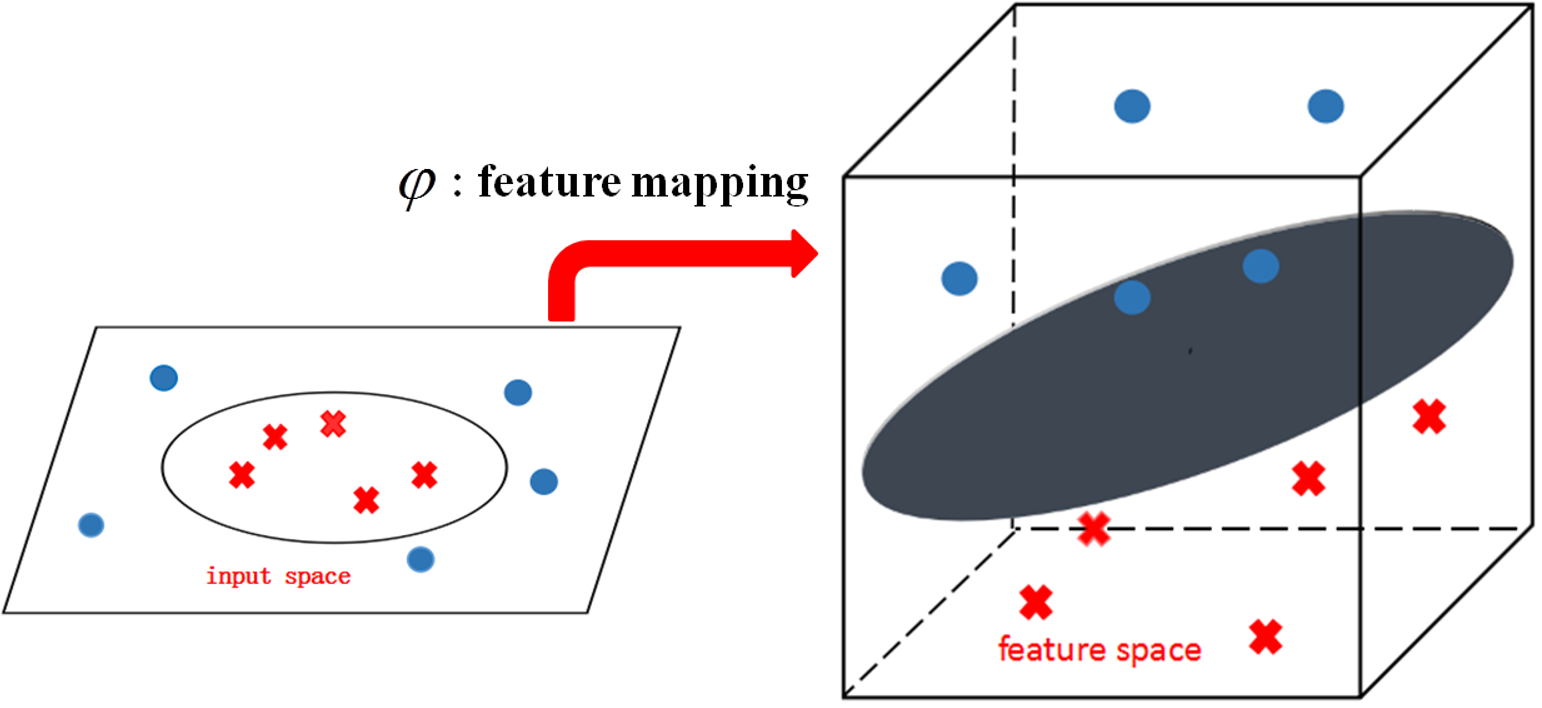}
%  \vspace{-.5cm}
%\centerline{{\small (a) Not linearly separable}}
\end{minipage}
\hfill
\caption{Feature mapping in SVM
 }
\label{fig:svm_feature}
\end{figure}

\begin{figure}[!t]
\begin{minipage}[b]{0.49\linewidth}
\centering
\includegraphics*[scale=0.2]{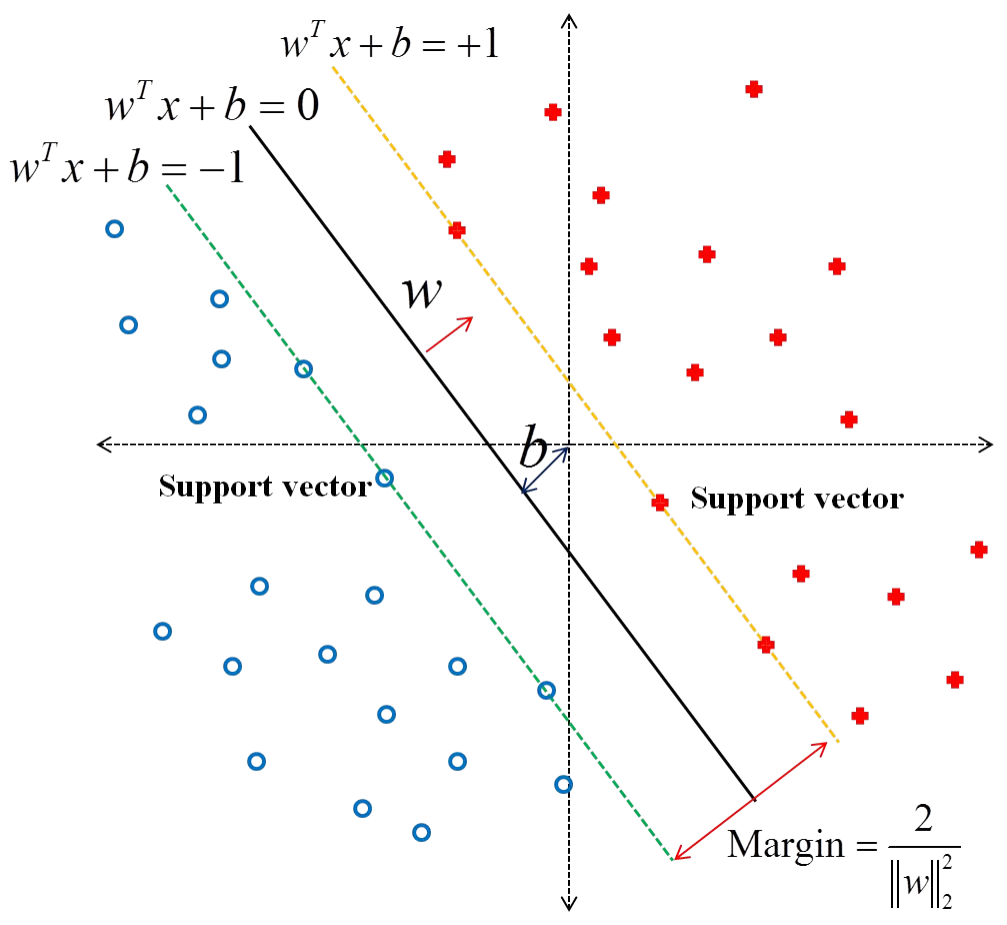}
%  \vspace{-.5cm}
\centerline{{\small (a) Margin principle for SVM}}
\end{minipage}
\hfill
\begin{minipage}[b]{0.49\linewidth}
\centering
\includegraphics*[scale=0.3]{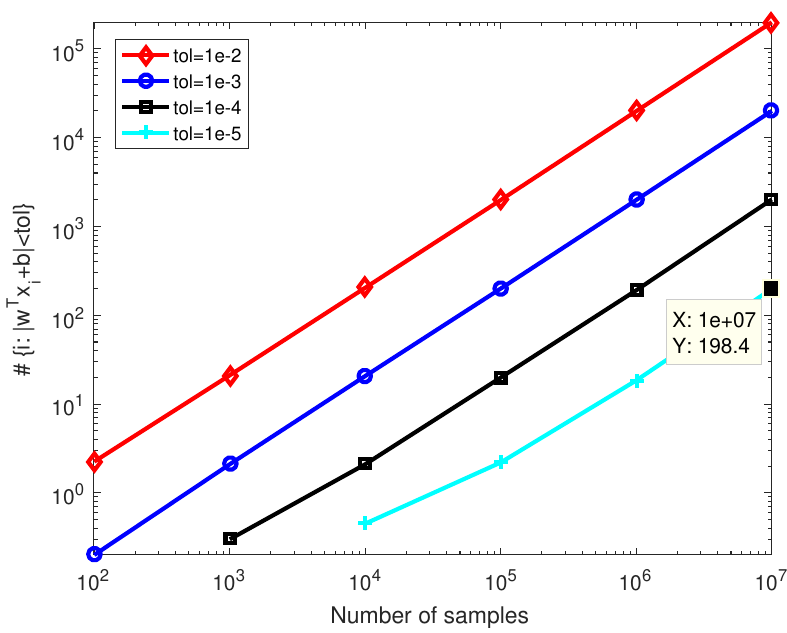}
%  \vspace{-.5cm}
\centerline{{\small (b) Margin v.s. data size}}
\end{minipage}
\hfill \caption{Maximal margin principle and its difficulty in massive data classification
 }
\label{fig:margin}
\end{figure}

SVM   is a classical and popular  method to tackle  binary classification problems.
The magic behind SVM is a feature mapping and a maximal margin principle.  As shown in Fig. \ref{fig:svm_feature}, the former  focuses on leveraging a low dimensional  input space into a high  dimensional feature space by some feature mapping such that the features associated with data are linearly separable.   The latter one, as exhibited in Figure \ref{fig:margin} (a), produces a unique classification decision by maximizing the margin  and then transforms the classification problem into a quadratic programming  problem.

Numerous practical applications  and theoretical studies \cite{Taylor2004,Steinwart2008} verified the power of SVM in classification, provided the size of data is not so large. However, when the size of data continuously increases, SVM is  confronted with two crucial design flaws. The first one, as shown in Fig. \ref{fig:margin} (b), is that the number of features  near the separation plane increases almost linearly with respect to the size of data, which makes the margin principle infeasible when the data size is huge. The other one is  that it requires at least $\mathcal O(m^2)$ memory requirements and usually $\mathcal O(m^3)$ computational complexity  to solve the  quadratic programming  problem, where $m$ is the number of data points.
%the size of data.
Both design flaws hinder  the use of SVM in tackling massive data classification problems.

\begin{figure}[!t]
\begin{minipage}[b]{0.49\linewidth}
\centering
\includegraphics*[scale=0.33]{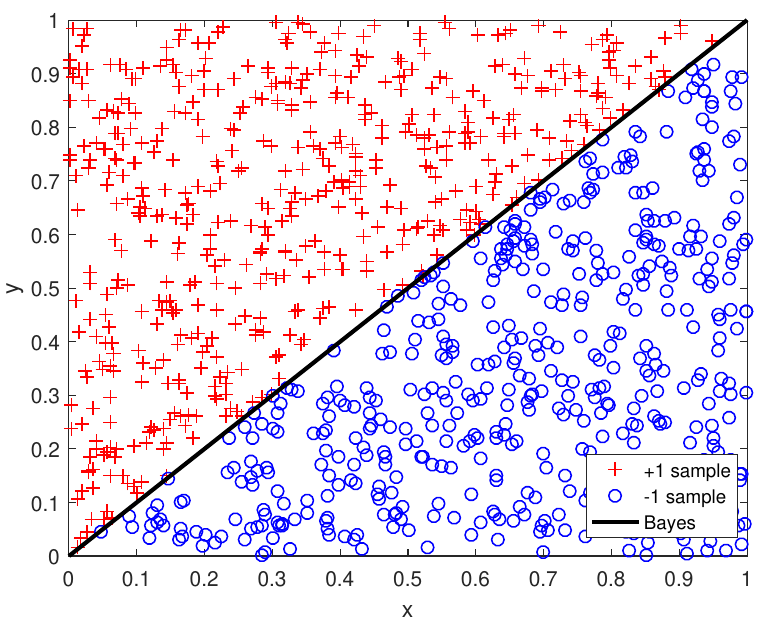}
%  \vspace{-.5cm}
\centerline{{\small (a) 2-dim separable without margin}}
\end{minipage}
\hfill
\begin{minipage}[b]{0.49\linewidth}
\centering
\includegraphics*[scale=0.33]{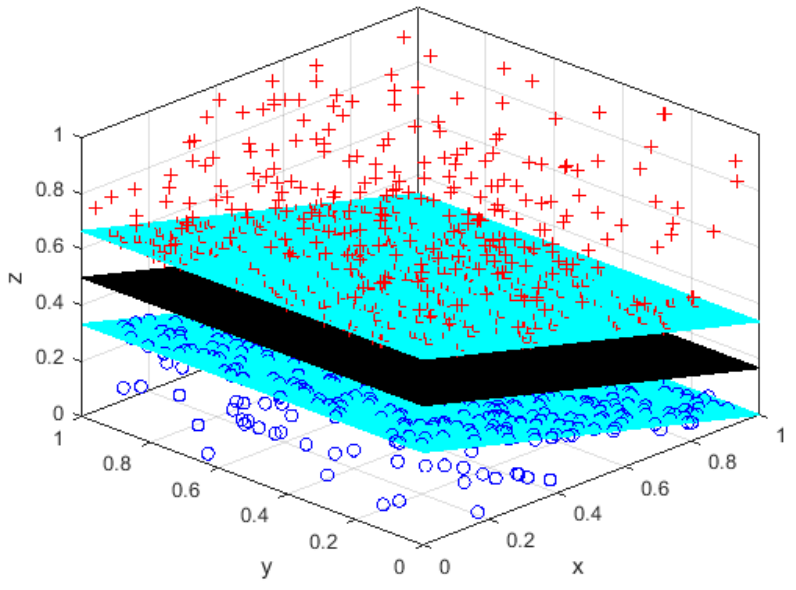}
%  \vspace{-.5cm}
\centerline{{\small (b) 3-dim separable}}
\end{minipage}
\hfill
 \caption{Leverage the dimension to guarantee the feasibility of the margin theory in the feature space
 }
\label{fig:effect-margin}
\end{figure}

Intuitively, there are two approaches to modify SVM to guarantee  the availability of the maximal margin principle. The first one is to leverage the dimension of the feature space further with additional feature mapping, just like Fig. \ref{fig:effect-margin} purports to show.  The other one is  the data reduction approach that  reduces the size of data  via selecting a small number of representative   samples. However, the former expands the feature space and thus needs more delicate algorithm to find a suitable classifier, which brings additional computational burden, while the performance of the latter depends heavily on  the representative   samples and usually requires clustering algorithms to find them out, which is also time-consuming for tackling massive data.

\begin{figure}[!t]
\centering
\includegraphics*[scale=0.5]{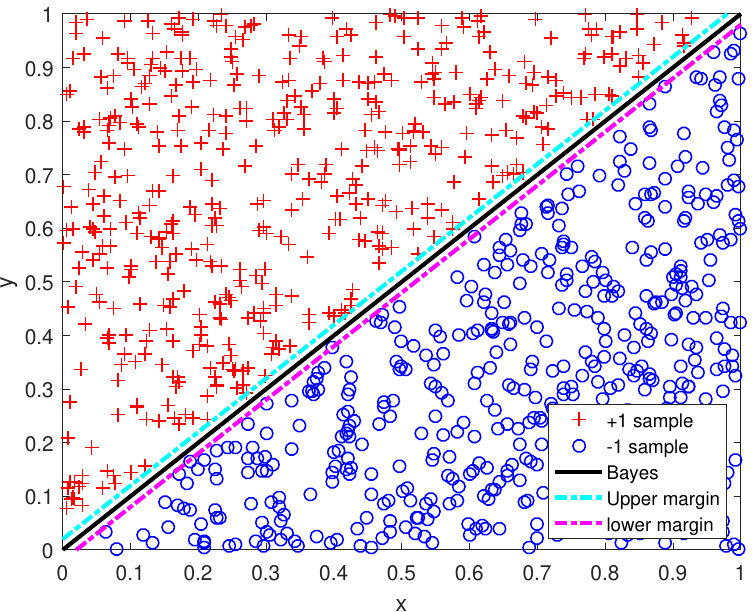}
%  \vspace{-.5cm}
 \caption{Margins for massive data
 }
\label{fig:motivations}
\end{figure}

In a nutshell, for massive data classification, it is time-consuming to produce  classifiers via the maximal margin principle. In this paper, we drive  a totally different direction from the maximal margin principle to design learning algorithms for massive data classification. Our basic idea is to select appropriate feature mappings such that   linear  classifiers with different margins in the feature space perform  similarly.  With these feature mappings,  the massiveness of data makes the margin of   support vectors be extremely small and thus it is not necessary to distinguish the linear classifier from the margin, just as  Fig. \ref{fig:motivations} exhibits.  With the maximal margin principle,  SVM is equivalent to solving a regularized  problem \cite{Taylor2004}
\begin{align}
\label{SVM}
f_{D,\lambda} = \arg\min_{f\in\mathcal H_K}\left\{\frac1m\sum_{i=1}^m(1-y_if(x_i))_++\lambda\|f\|_K^2\right\},
\end{align}
where
$D=\{(x_i,y_i)\}_{i=1}^{m}$ is the sample set, $t_+=\max\{t,0\}$ for $t\in\mathbb R$ and $\lambda$ is a regularization parameter which is proportional to the margin in Fig. \ref{fig:margin} (a). However, in our approach, since the maximal margin principle is not considered, we are faced with  an un-regularized convex optimization problem
\begin{align}
\label{Unregularzied-Motivation}
f_{D} = \arg\min_{f\in\mathcal H}\left\{\frac1m\sum_{i=1}^m(1-y_if(x_i))_+\right\},
\end{align}
where $\mathcal H$ is a linear space whose dimension is much smaller than $m$. Two important  ingredients in our approach are to find   suitable $\mathcal H$ and algorithms to solve (\ref{Unregularzied-Motivation}) such that (\ref{Unregularzied-Motivation}) performs at least as well as   (\ref{SVM}).

\subsection{Feature mapping with polynomial kernels}
\label{sc:featuremap}

In this subsection, we  focus on selecting the feature mapping for massive data classification.  Our purpose is to equip (\ref{Unregularzied-Motivation}) with a suitable feature space $\mathcal H$  such that (\ref{Unregularzied-Motivation}) performs similarly as (\ref{SVM}). Like SVM, we
constitute the feature mapping with kernels and the problem then boils down to choose kernels and centers for the kernel. As discussed in Subsection \ref{sc:motivations}, we are
desired for kernels which determine the dimension of the corresponding  feature space directly.  The most popular  kernel  for this purpose is the polynomial kernel $K_s(x,x')=(1+x\cdot x')^s$, where $s\in\mathbb N$ is the tunable parameter referring to the degree of  kernel polynomial and   $x,x'\in\mathbb R^d$.
Denote by
$\mathcal H_s$ the reproducing kernel Hilbert space (RKHS) associated with $K_s$ endowed with
an inner product $\langle\cdot,\cdot\rangle_s$
and norm
$\|\cdot\|_s$.
It is well known \cite{Zhou2006} that  ${\cal H}_s$ is the set of $d$-variable polynomials of degree at most $s$, and the dimension of ${\cal H}_s$ is $n = \left(^{s+d}_{\ d}\right) = \frac{(s+d)!}{d!s!}$.

Denote by $\mathcal P_s^d$ the set of algebraic polynomials defined on the input space  $X\subset\mathbb R^d$ of degree at most $s$.  Setting $\mathcal H=\mathcal P_s^d$, we are concerned with the
 empirical risk minimization:
\begin{equation}\label{KRR1}
f_{D} = \arg\min_{f\in\mathcal P_s^d}\left\{\frac1m\sum_{i=1}^m (1-y_if(x_i))_+\right\}.
\end{equation}
 Since the dimension of $\mathcal P_s^d$
is $ n=\left(^{s+d}_{\ s}\right)$,
if we select
$\{\eta_j\}_{j=1}^n\subset X$ such that $\{(1+\eta_j\cdot
x)^s\}_{j=1}^n$ is a linearly independent system, then
\begin{equation}
\label{hypotheisis n space}
\mathcal P_s^d=\left\{\sum_{j=1}^nc_j(1+\eta_j\cdot x)^s:c_j\in\mathbb R\right\}=:\mathcal H_{\eta,n}.
\end{equation}
In this way, \eqref{KRR1} can be converted to
\begin{equation}
\label{new model}
f_{D,n} = \arg\min_{f\in\mathcal H_{\eta,n}} \left\{\frac1m\sum_{i=1}^m (1-y_if(x_i))_+\right\}.
\end{equation}

To determine $\{\eta_j\}_{j=1}^n$, we  introduce the fundamental system with
respect to the polynomial kernel $K_s$ \cite{Lin2018-FPL} as follows.
Let $\zeta:=\{\zeta_i\}_{i=1}^n\subset X$. It is called a $K_s$-fundamental system if
\[
\mbox{dim}{\mathcal H_{\zeta,n}}=\left(_{\ s}^{s+d}\right),
\]
where $\mbox{dim}{\mathcal H}$ denotes the dimension of the linear space $\mathcal H$. It is easy to see that an arbitrary $K_s$-fundamental system implies \eqref{hypotheisis n space}.
The following proposition \cite{Lin2018-FPL} reveals that almost every set with $n=\left(_{\ s}^{d+s}\right)$ points is a $K_s$-fundamental system.

\begin{proposition}\label{Proposition:INDEPENDENCE}
Let $s,n\in \mathbb{N}$ and $n=\left(_{\ s}^{d+s}\right).$ Then the set
\[
\left\{\zeta=(\zeta_i)_{i=1}^n: \mathrm{dim}{\mathcal H_{\zeta,n}}<n\right\}
\]
has Lebesgue measure 0.
\end{proposition}

Based on Proposition  \ref{Proposition:INDEPENDENCE}, we can design simple strategies to choose the centers $\{\eta_j\}_{j=1}^n$.
In particular, $\{\eta_j\}_{j=1}^n$ can be selected either deterministically on $X$ or randomly independently and identically (i.i.d.) according to the uniform distribution,
since the uniform distribution is continuous with respect to the  Lebesgue measure.
 In summary,  there is only a discrete
parameter in (\ref{new model}) which reduces the difficulty of model selection.
Furthermore, since $n=\left(^{s+d}_{\ d}\right)$ and $s$ is frequently not larger than 10, $n$ is usually much smaller than $m$, especially when $d$ is not so large. Thus, the capacity of the feature space is usually small, reducing the difficulty for algorithm selection.

\subsection{ADMM for non-smooth convex optimization}
In this subsection, we try to  exploit the alternating direction method of multipliers (ADMM) \cite{Gabay1976} to solve the un-regularized optimization problem (\ref{new model}).
Let $A \in \mathbb{R}^{m\times n}$ with $A_{ij} = (1+ x_i\cdot  \eta_j)^s$.
To solve (\ref{new model}), it suffices to find a solution to    the following nonsmooth convex optimization problem:
\begin{align}
\label{Eq:opt_prob}
\min_{u\in \mathbb{R}^n} \frac{1}{m}\sum_{i=1}^m \left(1-y_i \sum_{j=1}^n A_{ij}u_j\right)_+.
\end{align}

Due to the nonsmoothness, the well-known gradient descent method is not available to the optimization problem \eqref{Eq:opt_prob}.
Concerning the sub-gradient  methods,  \cite{Nesterov2005} showed that   $\mathcal O(1/\varepsilon)$ iterations are required and   $\mathcal O(mn)$ float computations are needed in each iteration, where $\varepsilon$ is the approximation accuracy between the estimator generated by a sub-gradient method  and the global minimum of  (\ref{Eq:opt_prob}). Faced with massive data, $\varepsilon$ should be extremely small and thus sub-gradient methods involve extremely high computational burden.  From the optimization viewpoint, we can also develop some algorithms to (\ref{Eq:opt_prob}) based on its  dual form.  In the following proposition whose proof will be given in
%{\color{red} Appendix \ref{Appendix:Proof-Dual-Form}},
%\textit{Supplementary Material E},
\textit{Appendix E},
we present
  the dual form of   \eqref{Eq:opt_prob}.
\begin{proposition}
\label{Propos:dual-form}
The dual problem of \eqref{Eq:opt_prob} is a linear programming shown as follows:
\begin{align}
\label{Eq:dual-form}
&\max_{{\bf a, c}\in \mathbb{R}^m} \ {\bf 1}_m^T {\bf a}\\
&\mathrm{s.t.} \ {\bf a} + {\bf c} = \frac{1}{m}{\bf 1}_m, \ A^T \mathrm{Diag}(y) {\bf a}=0, \nonumber
\end{align}
where ${\bf 1}_m$ is the all one vector of dimension $m$, $\mathrm{Diag}(y)$ is a diagonal matrix with $y=(y_1,\dots,y_m)^T$ being its diagonal vector.
\end{proposition}

From Proposition \ref{Propos:dual-form}, the dual problem of the suggested learning scheme \eqref{Eq:opt_prob} is a linear program, which is  different from that of the classical SVM, i.e., the quadratic programming.
Noticing that the concerned problem is convex but nonsmooth, instead of sub-gradient methods,
we turn to ADMM, another type of powerful optimization methods that can handle the nonsmooth convex problem \eqref{Eq:opt_prob}.
We firstly reformulate the unconstrained problem \eqref{Eq:opt_prob} as the following equivalent constrained optimization problem via introducing another variable $v$,
\begin{align}
\label{Eq:opt_prob_admm}
&\mathop{\mathrm{min}}_{u\in \mathbb{R}^n, v\in \mathbb{R}^m} \ f(v) \quad \text{s.t.} \quad Au-v =0,
\end{align}
%\begin{align}
%\label{Eq:opt_prob_admm}
%&\mathop{\mathrm{min}}_{u\in \mathbb{R}^n, v\in \mathbb{R}^m} \ f(v)=\frac{1}{m}\sum_{i=1}^m \left(1-y_i v_i\right)_+ \ \ \text{s.t.} \ \ Au-v =0,
%\end{align}
where $f(v):=\frac{1}{m}\sum_{i=1}^m \left(1-y_i v_i\right)_+$.
The augmented Lagrangian function of (\ref{Eq:opt_prob_admm})  is defined by
\begin{align}
\label{Eq:ALM}
{\cal L}_{\beta}(u,v,w) = f(v) + \langle w, Au-v \rangle + \frac{\beta}{2} \|Au-v\|_2^2,
\end{align}
where $w\in \mathbb{R}^m$ is a multiplier variable, $\beta>0$ is the augmented Lagrangian parameter.
Based on ${\cal L}_{\beta}$, the ADMM algorithm for problem \eqref{Eq:opt_prob} can be described as follows:
given an initialization $u^0, v^0, w^0$, parameters $\alpha>0$, $\beta>0$, for $k=0,1,\ldots,$
\begin{align}
u^{k+1} & = \arg\min_{u\in \mathbb{R}^n} \ \left\{{\cal L}_{\beta}(u,v^{k},w^{k}) + \frac{\alpha}{2} \|u-u^{k}\|_2^2\right\},
      \label{Eq:update-u}\\
v^{k+1} & = \arg\min_{v\in \mathbb{R}^m} \ {\cal L}_{\beta}(u^{k+1},v,w^{k}),
    \label{Eq:update-v}\\
w^{k+1} & = w^{k} + \beta (Au^{k+1} - v^{k+1}). \label{Eq:update-w}
\end{align}
From \eqref{Eq:update-u}, we adopt the \textit{proximal} update strategy for $u^{k+1}$ instead of the original minimization strategy,
mainly due to the following two reasons:
the first one is to overcome the possible ill-conditionedness of matrix $A^TA$ via introducing a proximal term, as shown by \eqref{analytic-uk} in Lemma \ref{Lemm:solution-hinge-min} below,
and the second one is to stabilize the optimization procedure such that  successive two iterations change relatively smoothly.

\section{Fast Polynomial Kernel Classification}
\label{sc:feasibility}
In this section, we present the feasibility and parameter-selection of the ADMM algorithm for solving the non-smooth convex optimization problem  \eqref{Eq:opt_prob}. With these helps, we propose a novel algorithm called   fast polynomial kernel classification (FPC) for massive data clssification.

\subsection{On feasibility and convergence of ADMM}
To make ADMM user-friendly, we   present  closed-form solutions to  (\ref{Eq:update-u}) and (\ref{Eq:update-v}) such that  the sequence $u^k$, $v^{k}$ can be updated analytically.
 The following lemma shows the feasibility to solve the corresponding minimization problems.
\begin{lemma}
\label{Lemm:solution-hinge-min}
Let $(u^k, v^k, w^k)$ be the $k$-th iterate of ADMM.
Then the updates \eqref{Eq:update-u} and \eqref{Eq:update-v} can be expressed analytically as follows
%defined via (\eqref{Eq:update-u}) and (\ref{Eq:update-v}), respectively. Then, we get
\begin{equation}\label{analytic-uk}
     u^{k+1}= (\beta A^TA + \alpha {\bf I}_n)^{-1}(\alpha u^{k} + \beta A^T v^{k} - A^Tw^{k})
\end{equation}
and
\begin{equation}\label{analytic-vk}
      v^{k+1}
      = \mathrm{Hinge}_{m\beta}(y,Au^{k+1}+\beta^{-1}w^{k}),
\end{equation}
where ${\bf I}_n$ is the identity matrix of size $n$,
\begin{align*}
\mathrm{Hinge}_{\gamma}(\xi,\zeta)
= (\mathrm{hinge}_{\gamma}(\xi(1),\zeta(1)), \ldots, \mathrm{hinge}_{\gamma}(\xi(m),\zeta(m)))^T,
\end{align*}
$\xi=(\xi(1),\dots,\xi(m))^T$ for $\xi\in\mathbb R^m$, $\gamma>0$ and
\begin{align}\label{analytic-for-hinge}
&\mathrm{hinge}_{\gamma}(a,b) =\\
&\left\{
\begin{array}{cl}
b, &\ \mathrm{if} \ a=0,\nonumber\\
b+\gamma^{-1}a, &\ \mathrm{if} \ a \neq 0 \ \mathrm{and}\ ab\leq 1-\gamma^{-1}a^2,\nonumber\\
a^{-1}, &\ \mathrm{if} \ a \neq 0 \ \mathrm{and}\ 1-\gamma^{-1}a^2 <ab<1,\nonumber\\
b, &\ \mathrm{if} \ a \neq 0 \ \mathrm{and}\ ab\geq 1.\nonumber\\
\end{array}
\right.
\end{align}
\end{lemma}

The proof of this lemma will be presented in \textit{Appendix D}.
%\textit{Supplementary Material D}.
%{\color{red} Appendix D}.
 Lemma \ref{Lemm:solution-hinge-min} presents the feasibility of ADMM via showing  the closed-form update sequences  (\ref{Eq:update-u}), (\ref{Eq:update-v}) and (\ref{Eq:update-w}).
Besides the feasibility verification of (\ref{Eq:update-u}), (\ref{Eq:update-v}) and (\ref{Eq:update-w}), the update rules bring four types of parameters: $\alpha$ in (\ref{Eq:update-u}), $\beta$ in (\ref{Eq:update-u}), (\ref{Eq:update-v}), (\ref{Eq:update-w}), an initial point $(u^0,v^0,w^0)$ and a stopping rule. In the following theorem, we show that the proposed ADMM algorithm can get a global minimum of the non-smooth convex optimization problem  \eqref{Eq:opt_prob} and the convergence is independent of the selection of $\alpha,\beta$ and the initial point.

\begin{theorem}
\label{Thm:conv-admm}
Let $\{p^k:=(u^k,v^k,w^k)\}$ be the sequence generated by (\ref{Eq:update-u}), (\ref{Eq:update-v}) and (\ref{Eq:update-w}) for any $\alpha>0$, $\beta>0$ and finite initial point $(u^0,v^0,w^0)$.
Suppose that there exists a solution to problem \eqref{Eq:opt_prob}.
Then $p^k$ converges to some $p^*=(u^*,v^*,w^*)$ and $u^*$ is a global minimizer of \eqref{Eq:opt_prob}.
% \in \Omega^*
\end{theorem}

The proof of this theorem can be derived from \cite[Theorem 6.1]{He2015} and will be given
 in \textit{Appendix A} for completeness.
 %\textit{Supplementary Material A}.
% {\color{red} Appendix \Cref{Appendix:Proof-Conv-ADMM}}.
 Theorem \ref{Thm:conv-admm} shows the global convergence of the ADMM algorithm and presents theoretical    guidance on parameter-selection. In particular, since Theorem \ref{Thm:conv-admm} holds for any $\alpha,\beta>0$ and  finite initial point $(u^0,v^0,w^0)$, we can set $\alpha=\beta=1$ and $(u^0,v^0,w^0)=(0,y,0)$. In the following theorem, we present some guidance on setting the stopping rule.

\begin{theorem}
\label{Thm:conv-rate-admm}
Under the  assumptions of Theorem \ref{Thm:conv-admm}, we  have
\begin{align}
\label{Eq:monotone}
\|p^{k+1}-p^k\|_H^2 \leq \|p^k - p^{k-1}\|_H^2, \ \forall k\geq 1,
\end{align}
 and
\begin{align}
\label{Eq:conv-rate}
\|p^{k+1} - p^{k}\|_H^2 = o(1/k),
\end{align}
where
\begin{align}
\label{Eq:def-H}
H = \left( \begin{array}{ccc}
\alpha {\bf I}_n & 0                & 0\\
     0          &\beta {\bf I}_m   & 0\\
     0          &  0               & \beta^{-1}{\bf I}_m
\end{array}
\right),
\end{align}
and $\|\xi\|_H^2 = \xi^T H \xi$ for any $\xi \in \mathbb{R}^{2m+n}$.
\end{theorem}
The proof of Theorem \ref{Thm:conv-rate-admm} can be derived from \cite{He2015} and \cite{Deng2017-si}, and  will be presented in
\textit{Appendix B} for completeness.
Theorem \ref{Thm:conv-rate-admm} shows that the discrepancy between two successive iterations is monotonically decreasing at the rate of $o(1/k)$ under the metric of matrix norm.
This yields an efficient stopping criterion for ADMM, i.e., $\|p^{k+1} - p^{k}\|_H^2 < tol$ for some small positive tolerance constant $tol$.

\subsection{Fast polynomial kernel classification}
Based on the previous analysis, we aim at the non-smooth convex optimization problem \eqref{new model} and adopt the ADMM update rules (\ref{Eq:update-u}), (\ref{Eq:update-v}) and (\ref{Eq:update-w}) to generate a global minimum $\sum_{j=1}^nu^{*}_j K_s(\eta_j,\cdot)$
of \eqref{new model}, where $\{\eta_j\}_{j=1}^n$ is a $K_s$ fundamental system. With these, we propose a novel classification algorithm, called fast polynomial kernel classification (FPC),  for massive data classification.
\begin{algorithm}\caption{Fast Polynomial kernel Classification (FPC)}\label{alg1}
\begin{algorithmic}
\STATE {{\bf Input}:
training samples $D:=\{(x_i,y_i)\}_{i=1}^m$,  the degree $s\in\mathbb N$ of polynomial kernel $K_s(x,x')=(1+x\cdot x')^s$,  a $K_s$ fundamental system  $\{\eta_j\}_{j=1}^n$ of size $n$,
%with $n=\left(^{s+d}_{\ s}\right)$,
%$\alpha=\beta=1$, initialization $(u^0,v^0,w^0)=(0, y, 0)$, and stopping parameter $tol>0$
parameters $\alpha$ and $\beta$, initialization $(u^0,v^0,w^0)$ and stopping parameter $tol>0$ involved in ADMM}.
Let $A \in \mathbb{R}^{m\times n}$ with $A_{ij} = (1+ x_i\cdot  \eta_j)^s$.
%\STATE{{\bf Step 1}:
%Let $n=\left(^{s+d}_{\ s}\right)$ be the number of centers and $\{\eta_j\}_{j=1}^n$ be the set of centers, which is a $K_s$ fundamental            system.}
\STATE{{\bf Update:}
       For $k=0,1,\ldots,$ update}
\STATE{\quad  $u^{k+1}=(\beta A^TA + \alpha {\bf I}_n)^{-1}(\alpha u^{k} + \beta A^T v^{k} - A^Tw^{k})$,}
\STATE{\quad $ v^{k+1}
      = \mathrm{Hinge}_{m\beta}(y,Au^{k+1}+\beta^{-1}w^{k})$,}
\STATE{\quad $w^{k+1} = w^{k} + \beta (Au^{k+1} - v^{k+1})$.}
\STATE{{\bf End:} the first $k$ satisfying $\|p^{k+1} - p^{k}\|_H^2 < tol$ with $H$ defined by (\ref{Eq:def-H}).}
\STATE{{\bf Output}: $f_{D,s}(\cdot)=\sum_{j=1}^n u_j^{\mathrm{output}}K_s(\eta_j,\cdot)$.}
\end{algorithmic}
\end{algorithm}

As shown in Theorem  \ref{Thm:conv-admm}, we can use any  $\alpha,\beta$ and finite initialization $(u^0,v^0,w^0)$.
The default settings of these parameters are empirically chosen as $\alpha = 1$, $\beta = 1$ and $(u^0,v^0,w^0)=(0, y, 0)$ according to the experiment results later.
From the definition of the $K_s$  fundamental system and Proposition \ref{Proposition:INDEPENDENCE}, $\{\eta_j\}$ can be generated i.i.d. according to the uniform distribution from $X$ directly or from $\{x_i\}_{i=1}^m$, or drawn directly to be the first $n$ points from $\{x_i\}_{i=1}^m$. Our numerical results in Section \ref{sc:simulations} exhibit  that the selection of these parameters does not affect the performance of FPC very much.
For an iterative algorithm, an efficient stopping criterion is very important, especially when the size of data is huge. On one hand, Theorem \ref{Thm:conv-rate-admm} shows that   the distance  between two successive iterations, i.e., $\|p^{k+1}-p^k\|_H^2$, monotonically decreases to $0$ at the speed of $o(1/k)$, implying $tol$ the smaller the better. On the other hand, too small $tol$ will result in a large number of  iterations and thus requires huge computations.  Due to numerous practical trials,  we find that   $tol=5\times 10^{-4}$ is a good choice for massive data classification task. Of course, if the data size is not so large, we can set $tol$ to be extremely small to guarantee the convergence.
Therefore, there are only two parameters, the degree of kernel polynomial $s$ and number of selected centers $n$, to be tuned in FPC.
As shown in Theorem \ref{Thm:main} below, the optimal $s$ is less than  $\left(m/\log m \right)^{\frac{1}{d}}$.
If the input space has a relatively large dimension, then the optimal $s$ is generally less than $10$.
%, which makes $n$ be very small and thus reduces the computational and storage complexities of SVM.
Since $s$ is a discrete value, we use the well known cross-validation technique \cite[Chapter 8]{Gyorfy2002} to choose the best $s$ from the set $\{1,2,\ldots, 10\}$ in practice.
According to Proposition \ref{Proposition:INDEPENDENCE}, $n$ can be theoretically set to be $\left(^{s+d}_{\ s}\right)$ when $d$ is moderately large, and can be tuned by a hand-optimal way  in high-dimensional cases.
The FPC algorithm is summarized in Algorithm \ref{alg1}.

Next, we analyze the computational complexity of the proposed FPC.
Since $(\beta A^TA + \alpha {\bf I}_n)^{-1}$ is applied for each iteration, we  calculate and store it in advance, which requires $\mathcal O(mn)$    storage complexity and $\mathcal O(mn^2+n^3)$ computational complexity, respectively.
Once the inverse matrix is calculated in advance,
for each iteration,
the computational cost to update $u^{k+1}$ is ${\cal O}(mn + n^2)$.
By Lemma \ref{Lemm:solution-hinge-min}, $\mathrm{Hinge}_{m\beta}$ is an element-wise operator.
Thus, as shown by \eqref{analytic-for-hinge}, the computational cost of the update of $v^{k+1}$ is ${\cal O}(mn)$.
It is obvious that the computational cost of the update of $w^{k+1}$ is also ${\cal O}(mn)$.
Let $T$ be the maximal number of iterations achieving the given stopping criterion.
Then the total computational cost of the proposed FPC is ${\cal O}(mn^2 + Tmn)$.
As shown by our simulations to be presented later, it usually suffices to use very few iterations (about 5 iterations) to achieve the default stopping criterion with $tol=5\times 10^{-4}$.
Since $n$ and $T$ are generally much less than $m$, especially for a huge $m$,
the total computational complexity of FPC ${\cal O}(mn(T+n))$  is far lower than ${\cal O}(m^3)$ required for the classical SVM.
This shows that FPC is an efficient algorithm with an approximately linear computational complexity, and a good candidate for handling massive data classification tasks.

\subsection{Related works}

The existing variants of SVM for massive data classification   can be mainly divided into two categories.
The first class is  decomposition-based methods that divide  the original large scale quadratic programming (QP)  into smaller QP sub-problems \cite{Platt1999,Kao2004}.
Among these, the Sequential Minimal Optimization (SMO) algorithm proposed in \cite{Platt1999} is a representative one.
SMO transforms the large QP problem into a series of small QP problems, each involving only two dual variables according to the violation of the Karush-Kuhn-Tucker (KKT) conditions.
The major advantage of SMO is that each QP sub-problem can be solved analytically in an efficient way, without a numerical QP solver. Another advantage of SMO is that extra matrix storage is not required for keeping the kernel matrix, since no matrix operations are involved.
However, SMO converges slowly for massive data classification problem \cite{Kao2004}, mainly due to each iteration only involves two dual variables in the optimization. For example, when the size of training samples is in a million level, then the iteration number of SMO to run ergodically for all training samples is also  in a million level, which limits its application in massive data classification.

The second class is the data reduction-based methods which reduce  the number of training data points via selecting a small number of representative training samples from the large data set
 \cite{Chien2010,Tsang2005,Shen2016}.
%One sub-class of these methods are based on the subsampling schemes such as the random sampling, systematic sampling and importance sampling \cite{Lee2001,Chang2004,Chien2010,Tong2001,Tsang2005}.
%Another sub-class of these methods is based on the clustering \cite{Gu2013,Shen2016}.
Such a  method firstly discards many training samples according to either  clustering or sub-sampling schemes, and then implements SVM on the rest training samples.
Thus, its effectiveness  depends heavily  on the quality of selected training samples.
As a consequence, the effectiveness of sampling scheme and clustering scheme plays a central role for these methods, while the sampling or clustering scheme generally is associated to the a-prior information of the data distribution.
Moreover, discarding amount of training samples may result in  waste of data.
Although the practical effectiveness of these variants for massive data classification was verified, most of them  lack  theoretical guarantees on the generalization ability.

In the design flow of FPC, we propose a different direction to avoid the  maximal margin theory in classification.  Using the special features for polynomial kernels, we propose
%an ADMM rather than QP
an ADMM solver rather than a QP solver
to find the classifier. The advantages of FPC are the reduction of the tunable parameters, user-friendly design flow, low computational burden and optimal generalization error guarantee given in the following section.

The main reason of using polynomial kernels in FPC is that the capacity of the corresponding RKHS can be determined by the kernel parameter $s$. In this way, we can tune $s$ to balance the bias and variance. Our theoretical results in the next section and numerical results Sec.\ref{sc:simulations} illustrate  that removing the regularization term in SVM with polynomial kernels does not degenerate the learning performance. It should be mentioned that our results cannot be extended to SVM with other kernels such as the widely used Gaussian kernel, since the RKHS corresponding to a Gaussian kernel with an arbitrary kernel parameter (width) is of infinite dimension \cite{Minh2010}. It would be interesting to design scalable classification algorithms based on SVM with general kernels.

\section{Generalization Error Analysis}
\label{sc:theoretical-behavior}

Our analysis is  carried out in a standard binary classification framework \cite{Steinwart2008}, where
  the sample
$D=\{(x_i,y_i)\}_{i=1}^{m}$ with $x\in X$ and $y\in
Y=\{-1,1\}$ is assumed to be  drawn independently and identically according to  some  unknown
  distribution $\rho$, which admits a decomposition into the marginal distribution $\rho_X$ on $X$ and the conditional distribution $\rho(\cdot|x)$ at each $x\in X$. Binary
classification algorithms aim to generate  a classifier $\mathcal
C:X\rightarrow Y$, based on $D$,  to minimize the  misclassification error
$$
      \mathcal R(\mathcal C)=\mathbf P [\mathcal C(x)\neq y]=\int_X
      \mathbf P[y\neq\mathcal C(x)|x]d\rho_X,
$$
 where   $\mathbf P[y|x]$ is the conditional probability at $x\in X$.
 Theoretically, the
Bayes rule
$$
         f_c(x)=\left\{\begin{array}{cc}
         1, & \mbox{if}\ \xi(x)\geq 1/2,\\
         -1, & \mbox{otherwise }
         \end{array}\right.
$$
  minimizes the misclassification error, where
$ \xi(x)=\mathbf P[y=1|x]$ is the Bayes decision function. Since
$f_c$ is independent of the classifier $\mathcal C$, the performance
of the classifier $\mathcal C$ can be measured by the excess misclassification
error $\mathcal R(\mathcal C)-\mathcal R(f_c)$. Without loss of generality, we assume $X$ to be a simplex on $\mathbb{R}^d$, which is defined by
\[
X = \{x\in \mathbb{R}^d: x(i) \geq 0, 1\leq i \leq d, 1-|x| \geq 0\},
\]
where $x = (x(1), x(2), \ldots, x(d)) \in \mathbb{R}^d$, $|x| = \sum_{i=1}^d x(i)$.

To present the generalization error, we should introduce some assumptions
 on the data.  The first one is the Tsybakov noise condition \cite{Tsybakov2004} defined as follows.

\begin{definition}\label{Def:Tsybakov}
Let $0\leq q\leq\infty$. We say that $\rho$ satisfies the Tsybakov
noise condition with exponent $q$ if there exists a constant
$\hat{c}_q$ such that
\begin{equation}\label{Tsybakov noise}
      \rho_X(\{x\in X:|2 \xi(x)-1|\leq \hat{c}_qt\})\leq
      t^q,\qquad\forall t>0.
\end{equation}
\end{definition}

The Tsybakov noise condition  reflects the close extent of the margin from the hard margin to the soft margin.
Such a condition plays an important role in   reducing the variance of the relative loss with its expectation and has been adopted in \cite{Steinwart2007,Xiang2009,Tong2016,Lin2017-GaussSVM} to quantify  learning rates of SVM.

Different from the standard smoothness assumption \cite{Guo2017,Ying2017,Zhou2018}, the
  other condition in this paper is the geometric noise assumption proposed in \cite{Steinwart2007}.
 Denote by
\begin{equation}\label{definition of frho}
    f_\rho(x):=\arg\min_{t\in\mathbb{R}}\int_Y(1-yt)_+d\rho(y|x).
\end{equation}
It can be found in \cite{Zhang2004-si} that $f_\rho(x)=\mbox{sgn}(2 \xi(x)-1).$
 Write $X_{-1}:=\{x \in X: f_{\rho}(x) <0\}, X_1 := \{x\in X: f_{\rho}(x)>0\}$, and $X_0:= \{x\in X: f_{\rho}(x)=0\}$. We get a distance function $x \mapsto \tau_x$ by
\[
\tau_x :=
\left\{
\begin{array}{ll}
\mathrm{dist}(x,X_0\cup X_1), & \mathrm{if} \  x \in X_{-1},\\
\mathrm{dist}(x,X_0\cup X_{-1}), & \mathrm{if} \  x \in X_{1},\\
0, & \mathrm{otherwise},
\end{array}
\right.
\]
where $\mathrm{dist}(x,A)$ denotes the distance of $x$ to a set $A$ with respect to the Euclidean norm. It is easy to see that  $\tau_x$ quantitatively describes the distance between $x$ and the ``decision boundary''.
With this distance function, we can  define the following geometric noise condition
\cite[Definition 3]{Steinwart2007}.
% taken from \cite[Definition 2]{Tong2016}.

\begin{definition}
\label{Def:Geometric}
Let $\alpha>0$. We say that $\rho$ satisfies the geometric noise condition with exponent $\alpha$ if there exists a constant $c>0$ such that
\begin{align}
\label{Geometric_noise}
\int_X |f_{\rho}(x)|\exp\left(-\frac{\tau_x^2}{t}\right) d\rho_X \leq ct^{d\alpha/2}
\end{align}
holds for all $t>0$.
\end{definition}

The geometric noise condition describes the concentration of the measure $|f_{\rho}(x)|d \rho_X$ near the decision boundary and does not imply any smoothness of $f_c$ or regularity of $\rho_X$ with respect to the Lebesgue measure on $X$.
In particular, (\ref{Geometric_noise}) shows that the less $|f_{\rho}(x)|d \rho_X$  is concentrated near the decision boundary the larger the geometric exponent is. We refer the readers to \cite[Chap.2]{Steinwart2007} for more details  and examples of the geometric noise assumption. We highlight that  for the hinge loss the well known smoothness assumption    implies that $\{x\in
X:f_\rho(x)> 0\}$ and $\{x\in X:f_\rho(x)< 0\}$ have a strictly
positive distance, which is a bit strict \cite{Lin2017-GaussSVM} and the geometric noise condition is much more suitable.
In fact, the geometric noise condition can be guaranteed by a simple regularity condition on $f_\rho$ in terms that $|f_{\rho}(x)| \leq c_{\gamma}\tau_x^{\gamma}$ for some constants $\gamma$ and $c_{\gamma}$ and the Tsybakov noise exponent $q$ with a geometric noise exponent $\alpha = \frac{q+1}{2}\gamma$ if $q\geq 1$. Under this circumstance, if (\ref{Tsybakov noise}) holds with exponent $q\geq 1$, the geometric noise assumption with exponent $\alpha$   is equivalent to the simple regularity assumption $|f_{\rho}(x)| \leq c_{\gamma}\tau_x^{\gamma}$ with $\gamma=\frac{q+1}2$.
In the following theorem, we derive the generalization error bounds of FPC under    assumptions   \eqref{Tsybakov noise} and \eqref{Geometric_noise}.

\begin{theorem}
\label{Thm:main}
Assume that $\rho$ satisfies noise assumptions  \eqref{Tsybakov noise} and \eqref{Geometric_noise} with exponent $q>0$ and $\alpha>0$. For an arbitrary $0<p<2$, if  $s=\left[m^{\theta^*}\right]$ with $\theta^* = \frac{q+1}{\alpha(q+2+pq/2)+d(q+1)}$,  then for all $0< \delta <1,$ with confidence at least $1-\delta$, we have
\[
{\cal R}(\mbox{sgn}(f_{D,n})) - {\cal R}(f_c) \leq C m^{-\alpha \theta^*} \log (\frac{1}{\delta}),
\]
where $[a]$ represents the integer part of $a>0$ and $C$ is a positive constant independent of $m$ or $\delta$.
\end{theorem}

The proof of this theorem will be  provided in
\textit{Appendix C}.
%\textit{Supplementary Material C}.
Studying the generalization performance in the framework of learning theory is a classical topic in machine learning. In Steinwart's pioneer work \cite{Steinwart2001}, the   universal consistency of SVM with different kernels was presented. With this, \cite{Chen2004} provided generalization error bound for SVM with $q$-hinge loss under some smoothness assumptions for the Bayes decision function. For the polynomial kernel, following the work of \cite{Zhou2006,Tong2008}, \cite{Tong2016} proved that under the same condition as Theorem \ref{Thm:main}, SVM with polynomial kernels can reach a learning rate of  order  $m^{-\frac{\alpha(q+1)}{\alpha(q+2)+(d+1)(q+1)}}$.
Specifically,
ignoring  the extremely small term $qp/2$ in Theorem \ref{Thm:main}, the exponent of the learning rate of FPC is $\frac{\alpha(q+1)}{\alpha(q+2)+d(q+1)}$, which  is better than that of   SVM with polynomial kernels, i.e., $\frac{\alpha(q+1)}{\alpha (q+2)+(d+1)(q+1)}$ in \cite{Tong2016}. It is also comparable with SVM with Gaussian kernels in \cite{Steinwart2007}.
  If $\alpha=\infty$,  ignoring the extremely small term $qp/2$, the
learning rate  derived in Theorem \ref{Thm:main} is of order
$m^{-\frac{q+1}{q+2}}$. This rate coincides with the optimal
learning rates $m^{-\frac{q+1}{q+2}}$ for certain classifiers based
on empirical risk minimization in \cite{Tsybakov2004}.

\section{Toy simulations}
\label{sc:simulations}

In this section, we present  toy simulations to demonstrate the effect of algorithmic parameters of FPC and verify the developed theoretical assertions.
All the numerical experiments were carried out in Matlab R2015b environment
running Windows 10, Intel(R) Xeon(R) CPU E5-2667 v3 @ 3.2GHz 3.2GH, RAM 256GB. The code is available in the web site: \url{https://github.com/JinshanZeng/FPC}.

\subsection{Experimental setting}

The settings of simulations are described as follows.

\begin{figure}[!t]
\begin{minipage}[b]{0.98\linewidth}
\centering
\includegraphics*[scale=0.5]{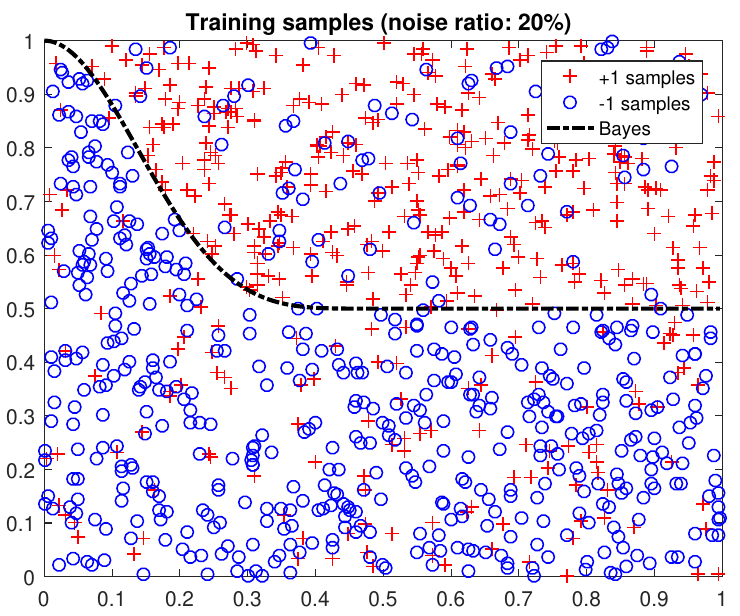}
%  \vspace{-.5cm}
%\centerline{{\small (a) Convergence rate (q=1)}}
\end{minipage}
\hfill \caption{ The generated data used in simulations. The red points are labeled as ``+1'' class, while the blue points are labeled as ``-1'' class.
} \label{fig:data}
\end{figure}

{\bf Samples}: In simulations, the training  samples were generated as follows.
Let
$$
   h(t)=\left((1-2t)_+^5(32t^2+10t+1)+1\right)/2,\qquad t\in[0,1]
$$
be a nonlinear Bayes rule.
Let ${\bf x}=\{x_i\}_{i=1}^m \subset ([0,1]\times [0,1])^m$
be drawn i.i.d. according to the uniform
distribution with size $m$.
Then we labeled the samples lying in the epigraph of function $h(t)$ as the positive class, while the rest were labeled as the negative class,
that is, given an $x_i = (x_i(1),x_i(2))$, its label $y_i =1$ if $x_i(2) \geq h(x_i(1))$, and $y_i=-1$ otherwise.
Moreover, for the training samples, we added $r\%$ noise, that is, we selected $m*r\%$ training samples via a uniformly random way and reversed their labels. Some training samples are depicted in Fig. \ref{fig:data}.
The testing samples ${\bf x}'=\{x_i'\}_{i=1}^{m'} \subset ([0,1]\times [0,1])^{m'}$ were generated according to the same procedure of training samples without adding noise.

{\bf Implementation and Evaluation}:
We implemented six simulations to verify the theoretical assessments and show the effectiveness of FPC.
For each simulation, we repeated $\ell \in \mathbb{N}$ times of experiments and recorded its training and test error, which are defined as the ratios of the number of wrong training (test) labels learned to the training (test) sample size, respectively.
The first one is to suggest an efficient stopping criterion of the suggested ADMM algorithm.
The second and third ones aim to show the sensitivity of the suggested ADMM to its  parameters including the proximal parameter $\alpha$ and Lagrangian parameter $\beta$, respectively.
%The fourth one is to verify one of our motivations in terms of studying the role of regularization parameter in the support vector machine (SVM).
The fourth one is to study the role of kernel parameter $s$ in FPC via comparing it with SVM using polynomial kernel (called \textit{SVM-Poly} for short).
The fifth one is to suggest some efficient center generation mechanisms in terms of the generalization ability,
and the final one is to show the robustness of the proposed learning scheme to different types of noise.

\subsection{Simulation results}

In this part, we report the experimental results and present some discussions.

{\bf Simulation 1: On stopping criterion.}
The initialization and stopping criterion are  crucial for the practical implementation of an iterative algorithm.
In this simulation, we aim to provide an effective initialization as well as a stopping criterion.
Specifically, we set $m=m'=1000$, $\ell=50$ and $r=10$.
As demonstrated by Theorem \ref{Thm:conv-rate-admm}, the sequence $\|p^{k+1}-p^k\|_H^2$ is monotonically decreasing.
We suggested using $\|p^{k+1}-p^k\|_H^2 < \textit{Tolerance}$ for some $\textit{Tolerance}>0$ as the stopping criteria.
Moreover, we suggested setting the initialization $p^0=(0, y , 0)$.
Under the above settings, in this simulation, we verified $25$ \textit{Tolerance}'s in the form of $10^{-\gamma}$, where $\gamma$ was taken as the equispaced points from $0$ to $5$ with the stepsize $0.2$.
The other parameters were set as
\[
s=9, \ n = \left(_{\ 9}^{11}\right) = 55, \ \alpha =1, \ \beta = 1.
\]
Centers $\{\eta_j\}_{j=1}^n$ were selected via a uniformly random way.
%as the first $n$ inputs of $\{x_i\}_{i=1}^m$.

\begin{figure}[!t]
\begin{minipage}[b]{0.49\linewidth}
\centering
\includegraphics*[scale=0.33]{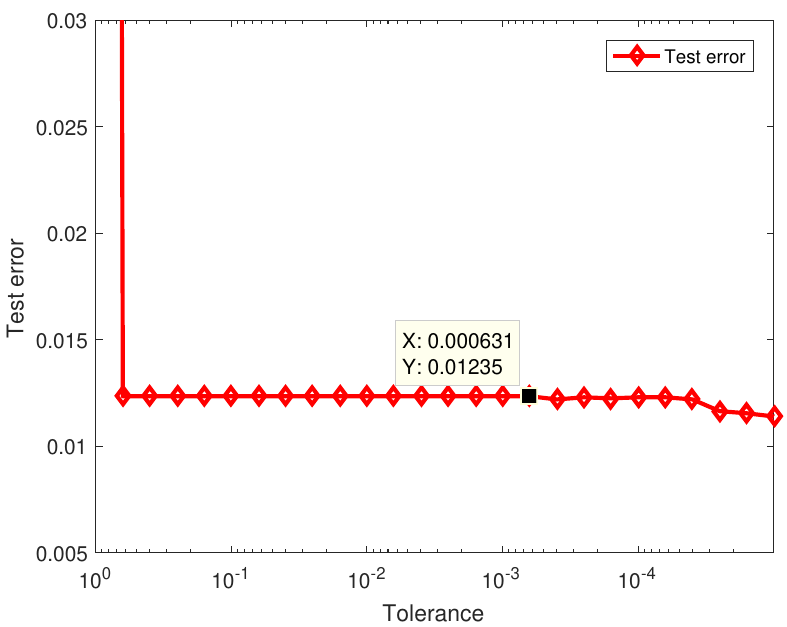}
%  \vspace{-.5cm}
\centerline{{\small (a) Test error}}
\end{minipage}
\hfill
\begin{minipage}[b]{0.49\linewidth}
\centering
\includegraphics*[scale=0.33]{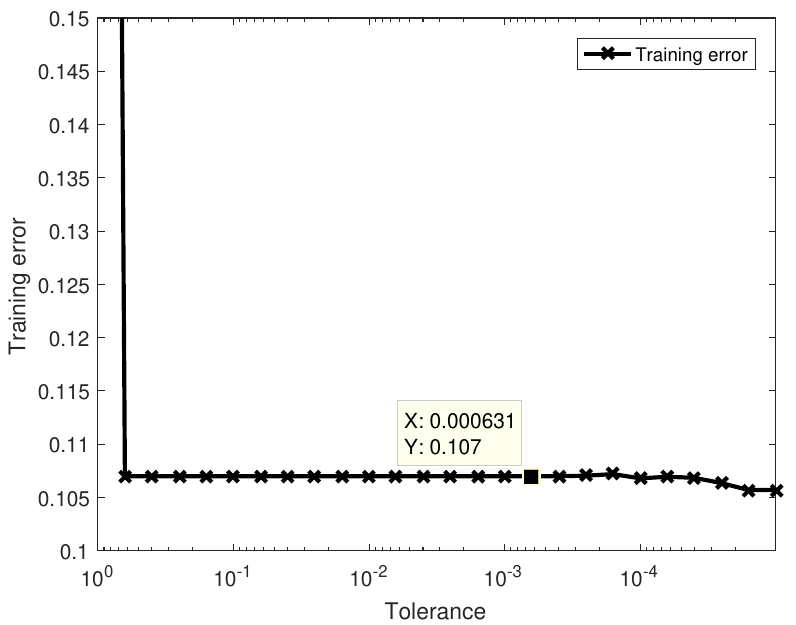}
%  \vspace{-.5cm}
\centerline{{\small (b) Training error}}
\end{minipage}
\hfill
\begin{minipage}[b]{0.49\linewidth}
\centering
\includegraphics*[scale=0.33]{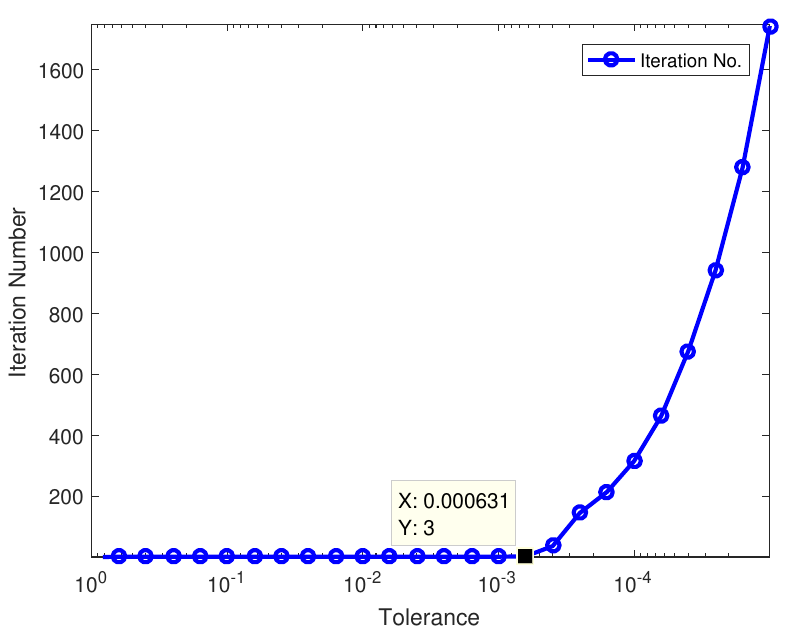}
%  \vspace{-.5cm}
\centerline{{\small (c) Maximal iteration no.}}
\end{minipage}
\hfill
\begin{minipage}[b]{0.49\linewidth}
\centering
\includegraphics*[scale=0.33]{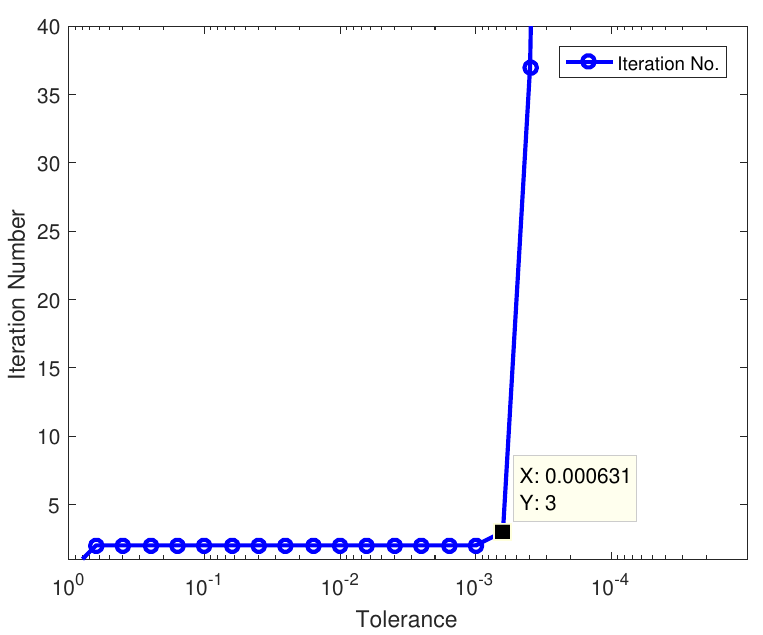}
%  \vspace{-.5cm}
\centerline{{\small (d) Detail of (c)}}
\end{minipage}
\hfill \caption{ The effect of stopping criterion for FPC.
 }
\label{fig:stopcrit}
\end{figure}

The trends of test error, training error, and the required maximal number of iterations are depicted in Fig. \ref{fig:stopcrit}.
From Fig. \ref{fig:stopcrit}, the test error of FPC is   stable for the choice of tolerance. However, when the tolerance is less than $10^{-3.2} (\approx 6.3 \times 10^{-4})$, the number of iterations required to achieve the given tolerance increases dramatically from $3$ to $1734$ as the tolerance decreases to $10^{-5}$, yet the test error only changes slightly from $0.01235$ to $0.01153$.
Therefore, in terms of both test error and computational cost, we suggest {\bf \textit{Tolerance}$=5\times 10^{-4}$} as the default stopping criterion of Algorithm \ref{alg1} in practice.

\begin{figure}[!t]
\begin{minipage}[b]{0.49\linewidth}
\centering
\includegraphics*[scale=0.33]{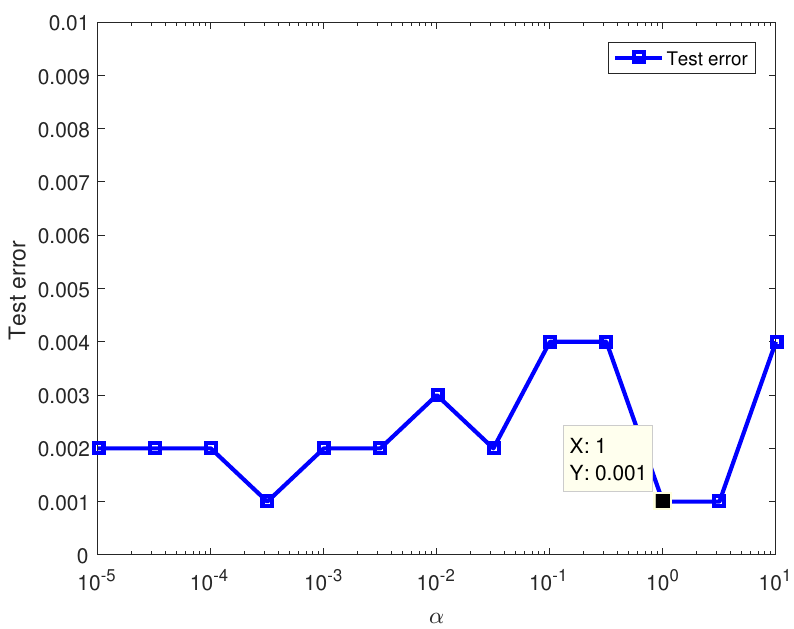}
%  \vspace{-.5cm}
\centerline{{\small (a) Test error}}
\end{minipage}
\hfill
\begin{minipage}[b]{0.49\linewidth}
\centering
\includegraphics*[scale=0.33]{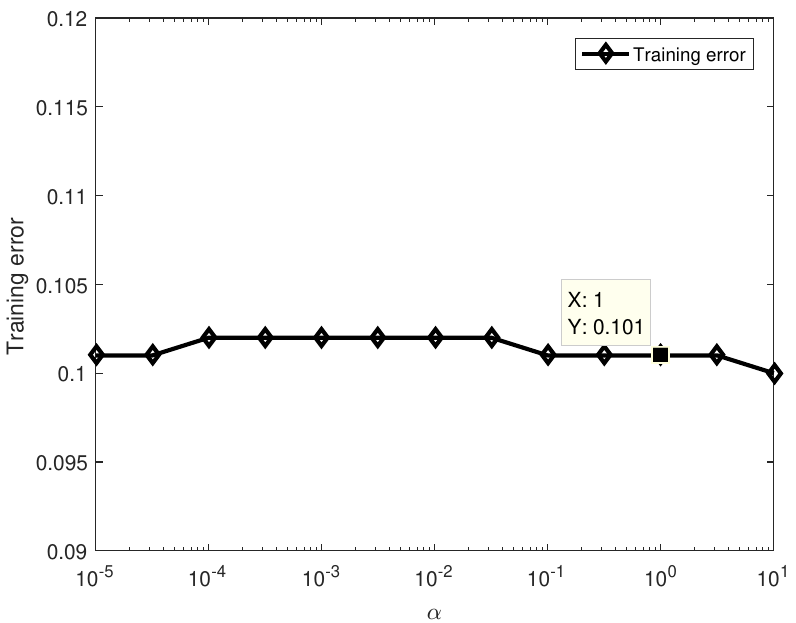}
%  \vspace{-.5cm}
\centerline{{\small (b) Training error}}
\end{minipage}
\hfill
\caption{ The effect of parameter $\alpha$ for FPC.
 }
\label{fig:effect_alpha}
\end{figure}

{\bf Simulation 2: On effect of proximal parameter $\alpha$.} In this simulation, we verified the effect of the proximal parameter $\alpha$ of the ADMM algorithm.
The settings of this simulation were similar to those in {\bf Simulation 1} except $\alpha$ varied in the form of $10^{\gamma}$,
where $\gamma$ was taken as the equispaced point of the interval $[-5,1]$ with the step $0.5$, and the stopping criterion was fixed as \textit{Tolerance}$=5\times 10^{-4}$.
The trends of test and training errors as the varying of $\alpha$ are presented in Fig. \ref{fig:effect_alpha}.
From Fig. \ref{fig:effect_alpha}, the proposed algorithm is stable for the choice of parameter $\alpha$.
Thus, we suggest using $\alpha =1$ as the default setting in practice for the computational convenience.

\begin{figure}[!t]
\begin{minipage}[b]{0.49\linewidth}
\centering
\includegraphics*[scale=0.33]{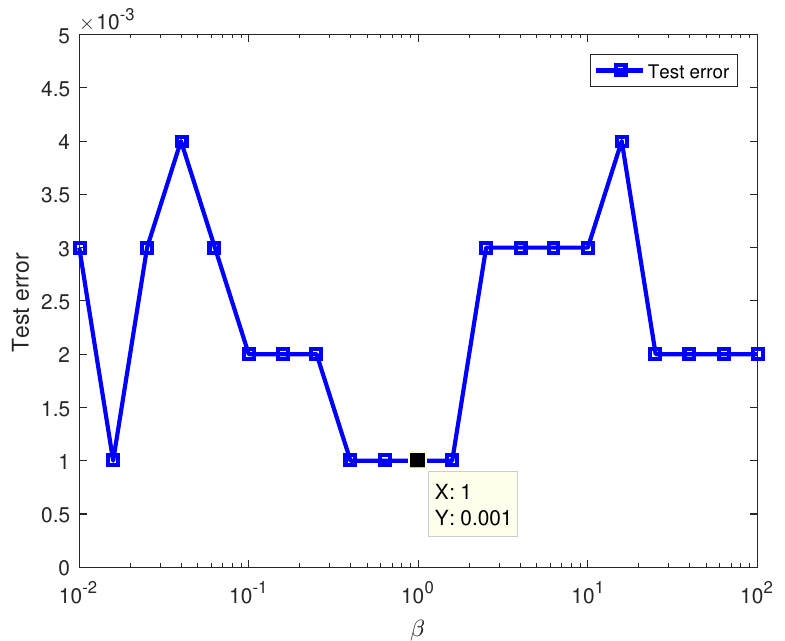}
%  \vspace{-.5cm}
\centerline{{\small (a) Test error}}
\end{minipage}
\hfill
\begin{minipage}[b]{0.49\linewidth}
\centering
\includegraphics*[scale=0.33]{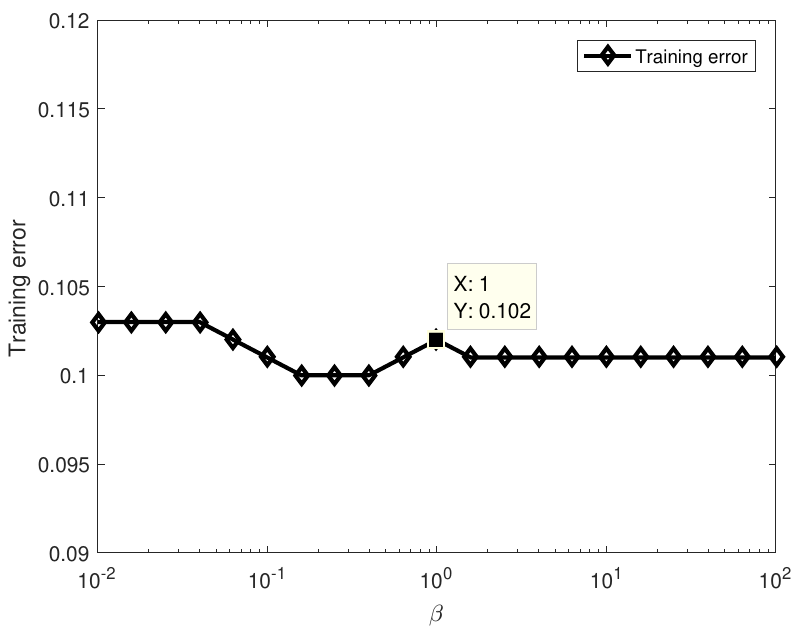}
%  \vspace{-.5cm}
\centerline{{\small (b) Training error}}
\end{minipage}
\hfill
\caption{ The effect of parameter $\beta$ for FPC.
 }
\label{fig:effect_beta}
\end{figure}

{\bf Simulation 3: On effect of parameter $\beta$.} As shown in Theorem \ref{Thm:conv-admm}, the suggested ADMM algorithm converges for arbitrary positive $\beta$.
In this simulation, we studied the numerical effect of the Lagrangian parameter $\beta$ of the ADMM algorithm.
The settings of this simulation were similar to those in {\bf Simulation 2} except $\beta$ varied in the form of $10^{\gamma}$ with
 $\gamma$ being taken as the equispaced point of the interval $[-2,2]$ with the step $0.2$, while $\alpha =1$ was fixed.
The trends of test and training errors as the varying of $\beta$ are presented in Fig. \ref{fig:effect_beta}.
From Fig. \ref{fig:effect_beta}, the choice of parameter $\beta$ has a little effect on the performance of FPC in terms of test error,
and the numerical performance of FPC is usually stable around $1$.
Thus, we suggest using $\beta =1$ as the default setting in practice.

\begin{figure}[!t]
\begin{minipage}[b]{0.98\linewidth}
\centering
\includegraphics*[scale=0.5]{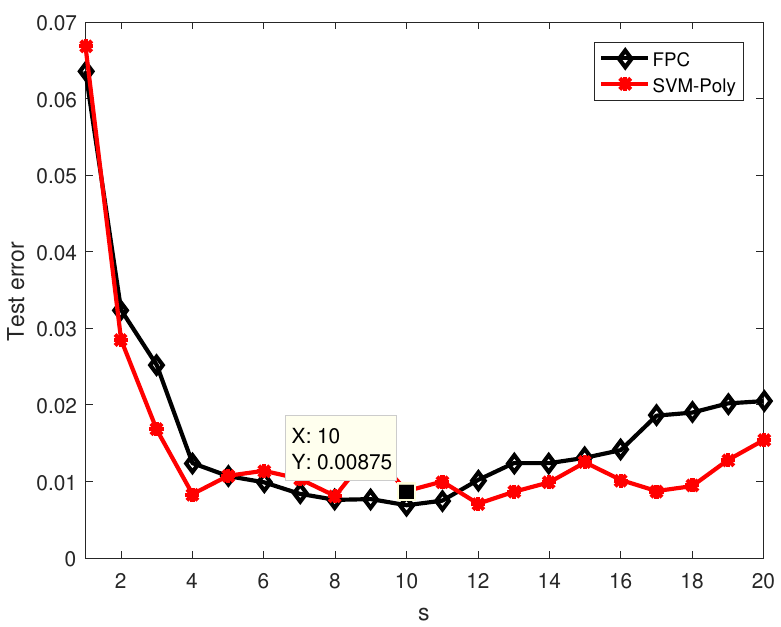}
\end{minipage}
\hfill \caption{ The effect of $s$ for both SVM-Poly (\ref{KRRp}) and FPC.
} \label{fig:effect_s}
\end{figure}

{\bf Simulation 4: On effect of kernel parameter $s$.}
In this simulation, we studied the importance of the parameter $s$ in SVM and FPC.
 We set $m=m'=1000$, $\ell=50$ and $r=10$. For comparison, we considered the following SVM with polynomial kernels (SVM-Poly)
\begin{align}\label{KRRp}
f_{D,s,\lambda}= \arg\min_{f\in\mathcal H_{s}}\left\{\frac1m\sum_{i=1}^m (1-y_if(x_i))_++\lambda\|f\|_s^2\right\}.
\end{align}
To solve \eqref{KRRp},
a quadratic programming (QP) of size $m$ should be solved via a numerical optimizer, such as SMO \cite{Platt1999} and \textit{libsvm} toolbox used in this paper.
The computational complexity of such a QP problem generally depends on the extent of ill-conditionedness of its associated coefficient matrix, i.e., $\mathcal K+\lambda m{\bf I}_{m}$, where $\mathcal K:=(K_s(x_i,x_j))_{i,j=1}^m$. In our running, the parameters of ADMM were set as follows:
$\alpha = 1$, $\beta = 1$, $Tolerance = 5\times 10^{-4}$ and $p^0 = (0,y,0)$.
Our aim is to  study whether there are  additional requirements for $s$ in FPC. For this purpose, we record  the test  errors of SVM and FPC with $s$ being varied from the range $[1,20]$.  For \eqref{KRRp}, $\lambda$ was selected to be optimal for a given $s$, according to the test samples directly.
We took test error as a function of $s$.
%by selecting the optimal $\lambda$ in \eqref{KRRp},
%from 20 candidates drawn equally spaced in the range $[1,20]$.
The simulation results are reported in Fig. \ref{fig:effect_s}.
By Fig. \ref{fig:effect_s}, there exist optimal degrees $s$ for  SVM-Poly and FPC, both of which are  around $10$.
Moreover,  trends of the effect of $s$ for both algorithms are  similar,  showing that the polynomial kernel parameter $s$ plays almost the same role and removing the regularization parameter in SVM does not bring additional difficulty in selecting $s$.

\begin{figure}[!t]
\begin{minipage}[b]{0.98\linewidth}
\centering
\includegraphics*[scale=0.5]{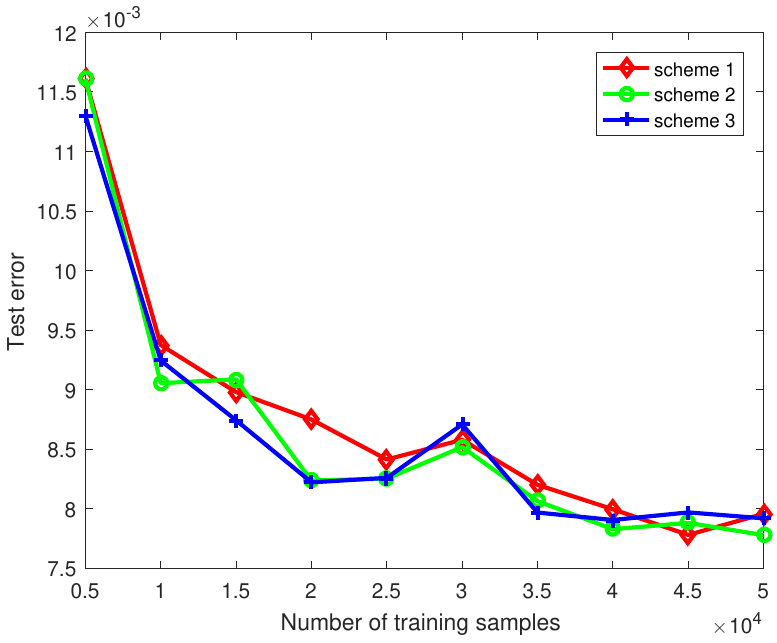}
\end{minipage}
\hfill
\caption{ The effect of   selection schemes for centers.
} \label{fig:comp_center}
\end{figure}

{\bf Simulation 5: On center generation mechanism.}
In this simulation, we set $m$ varying from $5\times 10^3$ to $5 \times 10^4$, $\ell=50$ and $r=10$.
Our aim is to study the effect of different center generation mechanisms for FPC.
Specifically, we considered the following three schemes for choosing $\eta$ in (\ref{Eq:opt_prob}).
Scheme 1 denotes that $\eta=\{\eta_i\}_{i=1}^n$ are drawn i.i.d according to the uniform distribution.
Scheme 2 denotes that $\{\eta_i\}_{i=1}^n$ are selected as the first $n$ inputs of samples.
Scheme 3 denotes that $\{\eta_i\}_{i=1}^n$ are selected randomly as the $n$ input of samples.
We figured out the test errors of these three approaches with different sizes of samples (from $5\times 10^3$ to $5 \times 10^4$) and optimal $s$ (selected according to the test samples).
The parameters of ADMM used in three schemes were the same as those in {\bf Simulation 4}.
The experimental results are reported in Fig. \ref{fig:comp_center}.
It can be found in Fig. \ref{fig:comp_center} that for a suitable $s$, the performance of three schemes are almost the same.
Thus, in practice, we usually suggest using \textit{Scheme 1} to generate the centers $\{\eta_j\}_{j=1}^n$ since it is data-independent.
%to generate the centers $\{\eta_j\}_{j=1}^n$, due to they are determined directly by $n$, and thus, there does not need additional storage to store them, once the training samples are given.

\begin{figure}[!t]
\begin{minipage}[b]{0.49\linewidth}
\centering
\includegraphics*[scale=0.33]{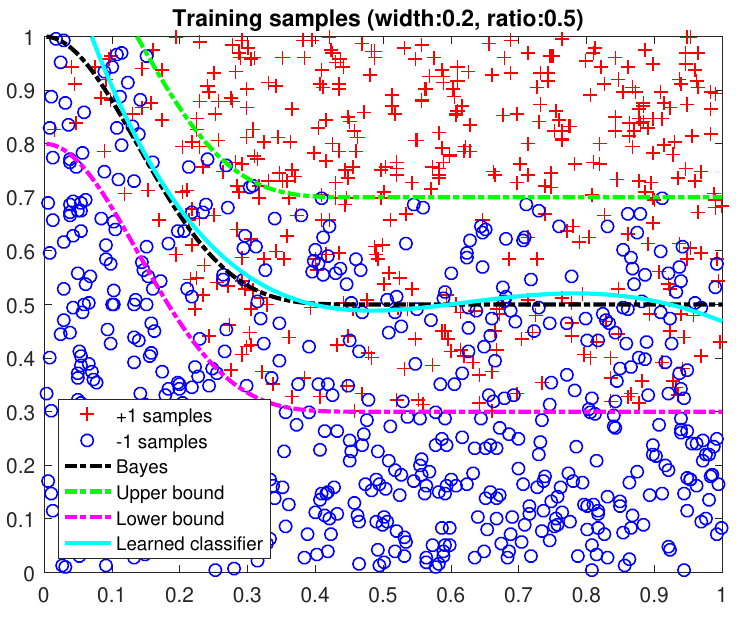}
%  \vspace{-.5cm}
\centerline{{\small (a) Classifier for noise type 1}}
\end{minipage}
\hfill
\begin{minipage}[b]{0.49\linewidth}
\centering
\includegraphics*[scale=0.33]{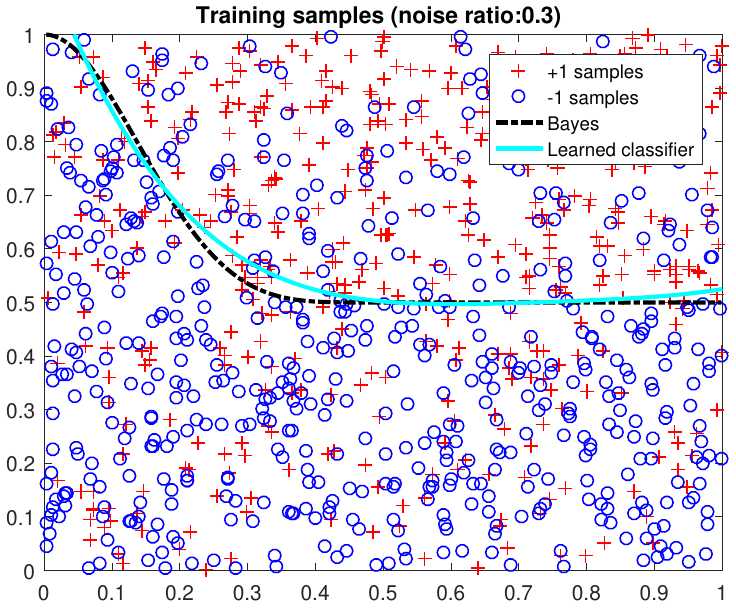}
%  \vspace{-.5cm}
\centerline{{\small (b) Classifier for noise type 3}}
\end{minipage}
\hfill
\begin{minipage}[b]{0.49\linewidth}
\centering
\includegraphics*[scale=0.33]{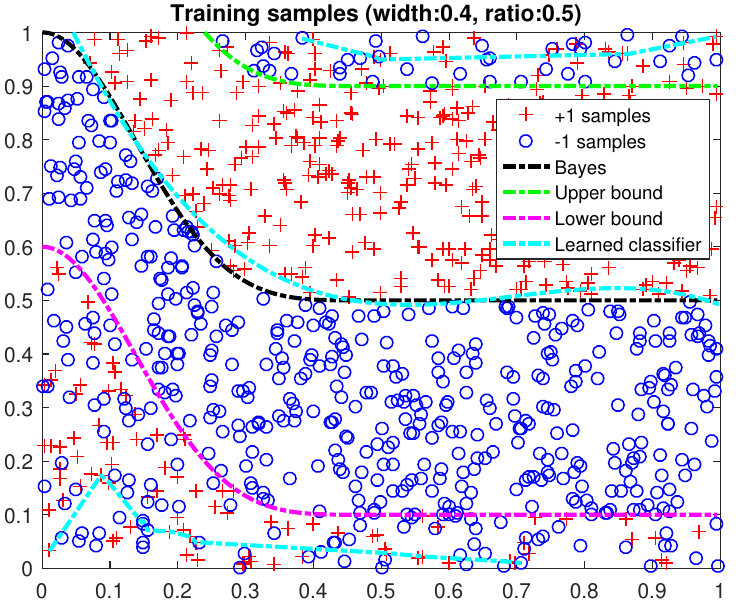}
%  \vspace{-.5cm}
\centerline{{\small (c) Classifier for noise type 2}}
\end{minipage}
\hfill
\begin{minipage}[b]{0.49\linewidth}
\centering
\includegraphics*[scale=0.33]{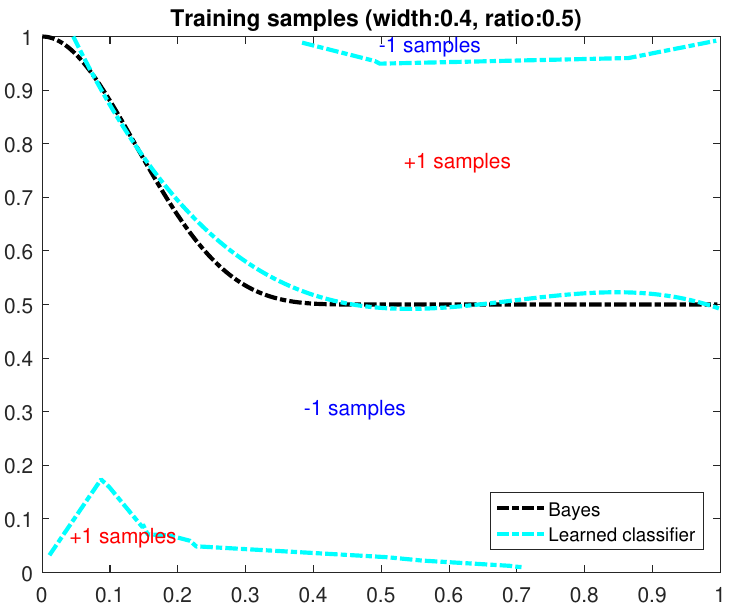}
%  \vspace{-.5cm}
\centerline{{\small (d) Detail of (c)}}
\end{minipage}
\hfill
\caption{ The classifiers learned by FPC under different types of noise, where the cyan lines are the learned classifiers of FPC, and the black one is the ground truth of Bayes.
For the first two types of noise, the \textbf{width} is defined as the distance between the boundaries of noise and Bayes, that is, the distance between the magenta and black lines for noise type 1, or the distance between the green and black lines for noise type 2, respectively,
and the \textbf{ratio} is defined as the ratio of noisy samples in the specified regions, that is, the region nearby the Bayes for noise type 1 and the region away form the Bayes for noise type 2, in (a) and (c) of this figure.
The \textbf{noise ratio} for the third type of noise is defined as the number of noisy samples with ``wrong'' labels divided by the total number of training samples.
 }
\label{fig:classifier}
\end{figure}

\begin{figure}[!t]
\begin{minipage}[b]{0.49\linewidth}
\centering
\includegraphics*[scale=0.33]{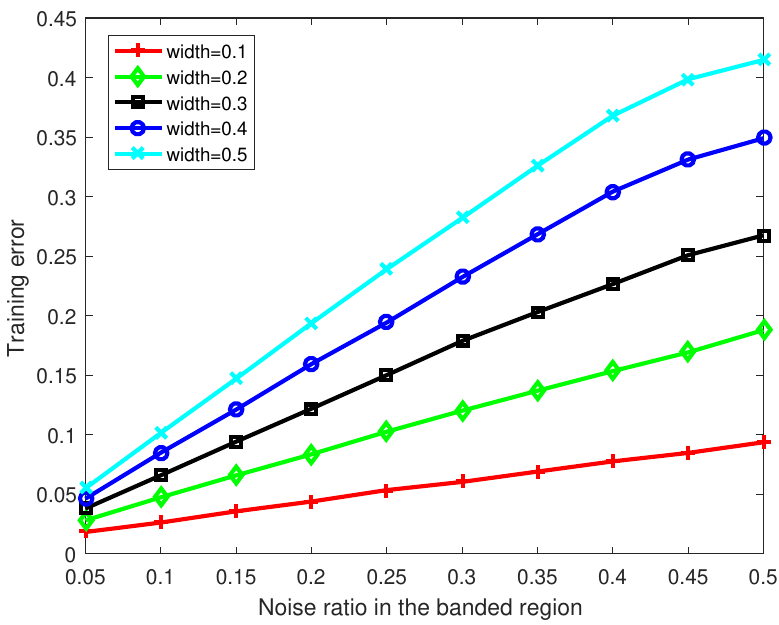}
%  \vspace{-.5cm}
\centerline{{\small (a) TrainErr for noise type 1}}
\end{minipage}
\hfill
\begin{minipage}[b]{0.49\linewidth}
\centering
\includegraphics*[scale=0.33]{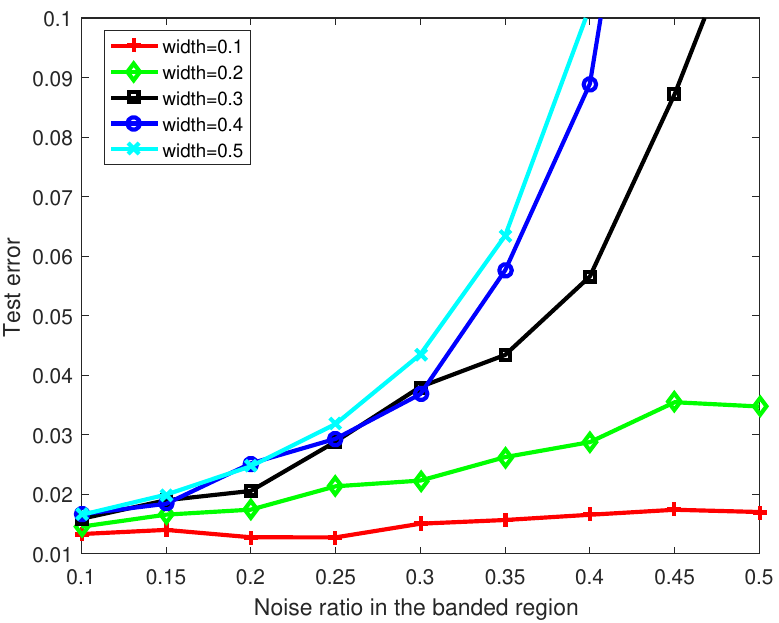}
%  \vspace{-.5cm}
\centerline{{\small (b) TestErr for noise type 1}}
\end{minipage}
\hfill
\begin{minipage}[b]{0.49\linewidth}
\centering
\includegraphics*[scale=0.33]{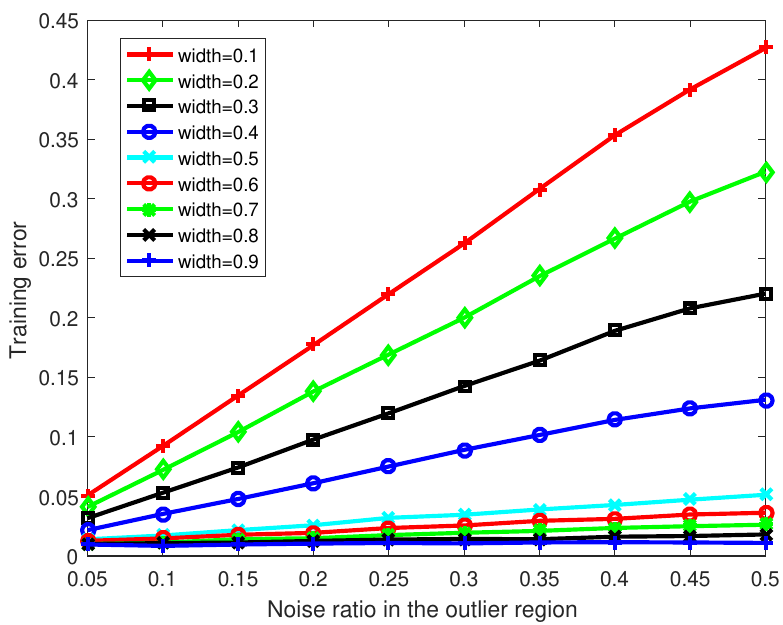}
%  \vspace{-.5cm}
\centerline{{\small (c) TrainErr for noise type 2}}
\end{minipage}
\hfill
\begin{minipage}[b]{0.49\linewidth}
\centering
\includegraphics*[scale=0.33]{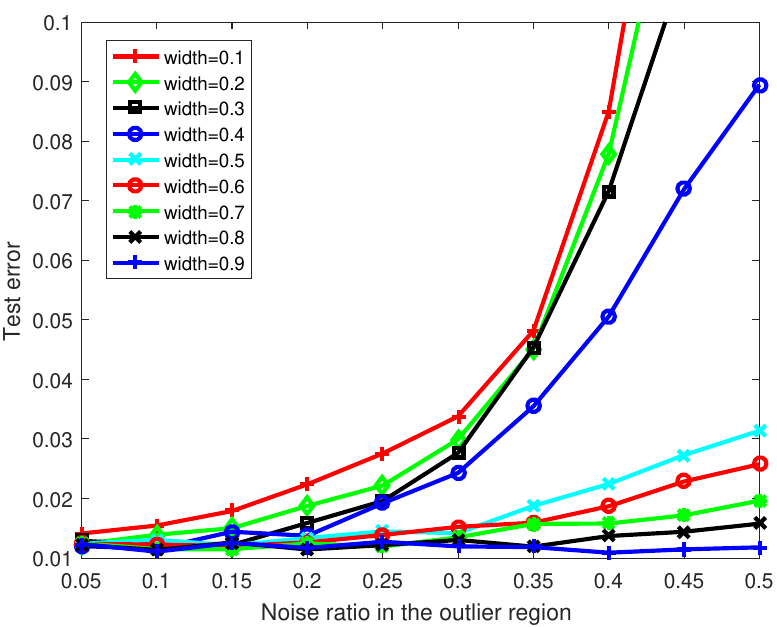}
%  \vspace{-.5cm}
\centerline{{\small (d) TestErr for noise type 2}}
\end{minipage}
\hfill
\begin{minipage}[b]{0.49\linewidth}
\centering
\includegraphics*[scale=0.33]{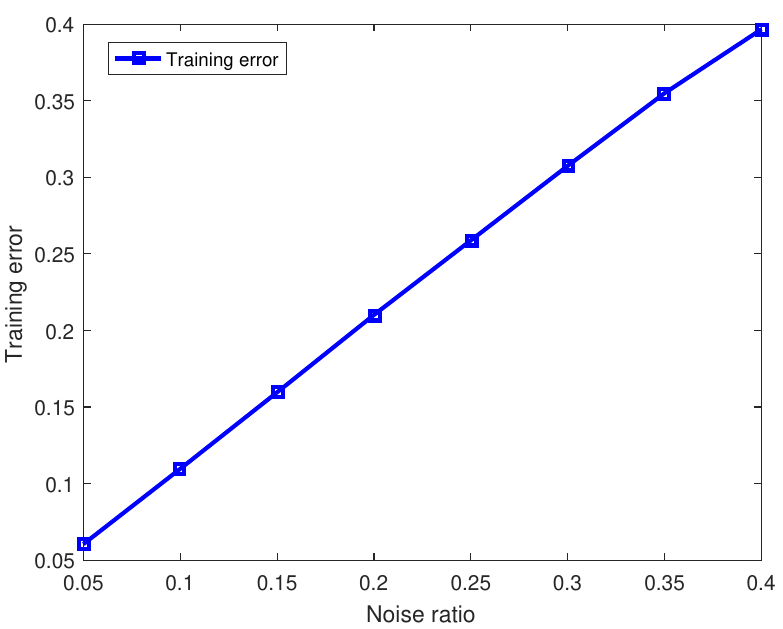}
%  \vspace{-.5cm}
\centerline{{\small (e) TrainErr for noise type 3}}
\end{minipage}
\hfill
\begin{minipage}[b]{0.49\linewidth}
\centering
\includegraphics*[scale=0.33]{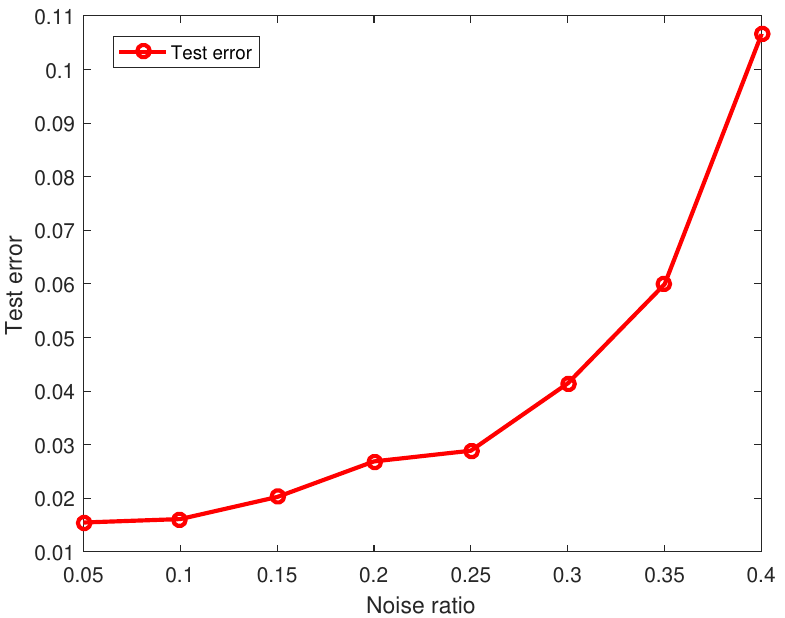}
%  \vspace{-.5cm}
\centerline{{\small (f) TestErr for noise type 3}}
\end{minipage}
\hfill
\caption{ Performance of FPC for three types of noise with differen levels.
 }
\label{fig:noise}
\end{figure}

{\bf Simulation 6: On effect of noise.}
In this simulation, we considered the effect of three different types of noise: the noisy samples concentrated within a band of the Bayes (see, Fig. \ref{fig:classifier}(a)); the noisy samples lying in the region that is far from the Bayes (see, Fig. \ref{fig:classifier}(c)); and the noisy samples lying randomly in all the region $[0,1]\times [0,1]$ (see, Fig. \ref{fig:classifier}(b)).
In the following, we considered different levels for each noise type.
For noise type 1 and type 2, we consider different widths of the banded region and noise ratio  within the banded region.
%The trends of training and test error are depicted in Figure \ref{fig:noise} (a) - (d).
%Similarly, we consider different widths and noise ratios for noise type 2, with the trends of training and test error being presented in Figure \ref{fig:noise} (c) and (d).
For noise type 3, we only considered different noise levels.
For all these three types of noise, the parameters of FPC were set the same as the following:
$\alpha = 1$, $\beta = 1$, $Tolerance = 5\times 10^{-4}$, $p^0 = (0,y,0)$, $s=9$, $n = \left(^{s+d}_{\ s}\right) =55$,
and the centers $\{\eta_i\}_{i=1}^n$ were set according to \textit{Scheme 1} as suggested in \textbf{Simulation 5}.
The trends of training and test errors with respect to the noise ratio are depicted in Fig. \ref{fig:noise} (a)-(f).
Particularly, the noise ratios  for noise type 1 and type 2 are defined as the ratio of the number of noisy samples to the size of training samples.
For noise type 1, the noise level roughly equals to the multiplication of the band width and the specified noise ratio in the considered region,
while for noise type 2, the noise level roughly equals to the multiplication of one minus the band width and the specified noise ratio in the considered region.

From Fig. \ref{fig:noise}, the training error of FPC is almost equal to the noise level, which means that FPC preserves these three different types of noise approximately linearly.
Thus, FPC is generally   robust to these types of noise.
By Fig. \ref{fig:noise} (b), (d) and (f), on  one hand, it is expected that the trend of test error gets worse as the increasing of noise level.
On the other hand, given a level of training error (which approximately reflects the noise level), FPC generally performs the best for noise type 3, while similarly for the other two types of noise.
Some learned classifiers by FPC for three different types of noise are depicted in Fig. \ref{fig:classifier}.
From Fig. \ref{fig:classifier}(a) and (b), the learned classifiers for the first and third types of noise are still preserved well under some moderate noise levels, as the learned classifiers are nearby the Bayes and keep the similar trends as the Bayes.
However, for the second type of noise, the learned classifier is generally divided into three pieces, that is, the main piece still nearby the Bayes, the other two pieces lie in the two outlier regions, as shown in Fig. \ref{fig:classifier}.
Such phenomenon is reasonable because the second type of noise actually cannot be simply regarded as   random noise, but it will be more suitable to be considered as some outliers, which changes the classifiers.

\begin{table}
\caption{Details of UCI data sets. The last column presents the percentage of the majority class of this data set. In the later tables, we use the first vocabulary of the name of the data set for short.}
\begin{center}
% \scriptsize
\begin{tabular}{|l|c|c|c|c|}\hline
  Data sets & Data size    & \#Attributes  & Majority (\%)\\\hline
  breast\_cancer           &683    & 9     & 64.44\\ \hline
  banknote\_authentication &1,372   & 4    & 55.54\\ \hline
  seismic\_bumps           &2,584   & 18   & 93.42\\ \hline
  musk2                    &6,598   & 166  & 84.59 \\ \hline
  HTRU2                    &17,898  & 8    & 90.84\\ \hline
  MAGIC\_Gamma\_Telescope  &19,020  & 10   & 64.84\\ \hline
  occupancy                &20,560  & 5    & 76.90\\ \hline
  default\_of\_credit\_card\_clients & 30,000  & 24  & 77.88\\ \hline
  {Skin\_NonSkin}          &245,057 & 3    & 79.25\\ \hline
\end{tabular}
\end{center}
 \label{Tab:uci-data}
\end{table}

\begin{table*}
\caption{Comparison results between the proposed method and existing methods on UCI data sets in terms of testing accuracy (\%). The best results are marked in bold.}
\begin{center}
% \tiny
\begin{tabular}{|c|c|c|c|c|c|c|c|c|c|}\hline
%   \multirow{2}{*}
   {Data set}
%   &
%\multicolumn{5}{|c|}{TestAcc (std.) \%} \\
%\cline{2-6}
                   & breast           & banknote        & seismic          & musk2             & HTRU2             & MAGIC             & occupancy         & default           & Skin             \\\hline
 SVM-RBF           & 97.19$\pm$0.71   & 98.07$\pm$0.71  & 93.84$\pm$0.37   & 91.11$\pm$0.41    & 97.53$\pm$0.06    & 85.69$\pm$0.26    & 98.63$\pm$0.08    & 81.60$\pm$1.05    & 98.80$\pm$0.02   \\\hline
 SVM-Poly          & 96.84$\pm$1.05   & 97.72$\pm$0.99  & 93.59$\pm$0.74   & 92.82$\pm$0.32    & 97.42$\pm$0.08    & 86.00$\pm$0.11    & 98.95$\pm$0.08    & 82.10$\pm$0.32    & 99.06$\pm$0.01   \\\hline
 linear-SVM        & 96.38$\pm$0.82   & 97.43$\pm$0.77  & 93.42$\pm$0.62   & 91.05$\pm$0.48    & 97.32$\pm$0.08    & 85.44$\pm$0.28    & 98.46$\pm$0.08    & 81.42$\pm$0.96    & 98.53$\pm$0.04   \\\hline
 $\ell_1$-SVM      & 96.73$\pm$0.78   & 97.84$\pm$0.81  & 93.62$\pm$0.56   & 92.44$\pm$0.46    & 97.56$\pm$0.08    & 85.83$\pm$0.18    & 98.55$\pm$0.06    & 81.80$\pm$0.64    & 98.77$\pm$0.05   \\\hline
 RF                & 96.81$\pm$1.40   & 98.99$\pm$0.57  & 92.88$\pm$0.78   & 95.56$\pm$0.51    & 97.88$\pm$0.18    &  86.90$\pm$0.53    & {\bf 99.14}$\pm$0.10    & 81.01$\pm$0.40    & {\bf 99.94}$\pm$0.01   \\\hline
 PROX-PFGM          & {\bf 97.24}$\pm$0.82   & 98.62$\pm$0.46  & 93.86$\pm$0.42   & 95.12$\pm$0.66    & 97.58$\pm$0.06    & 86.34$\pm$0.41    & 98.71$\pm$0.07    & {\bf 82.32}$\pm$0.43    & 99.12$\pm$0.02   \\\hline
 FALKON            & 96.93$\pm$0.88   & 98.74$\pm$0.62  & 93.48$\pm$0.65   & 95.46$\pm$0.36    & {\bf 97.93}$\pm$0.08    & {\bf 86.92}$\pm$0.44    & 98.74$\pm$0.08    & 82.09$\pm$0.50    & 98.84$\pm$0.02   \\\hline
 RR-Fourier        & 96.82$\pm$0.69   & 98.64$\pm$0.54  & 93.75$\pm$0.75   & 94.21$\pm$0.52    & 97.24$\pm$0.12    & 86.09$\pm$0.54    & 98.33$\pm$0.09    & 81.85$\pm$0.64    & 98.46$\pm$0.03   \\\hline
 TS                & 96.64$\pm$0.77   & 98.43$\pm$0.66  & 93.46$\pm$0.68   & 94.61$\pm$0.36    & 97.26$\pm$0.10    & 85.69$\pm$0.46    & 98.26$\pm$0.07    & 81.63$\pm$0.46    & 98.13$\pm$0.04   \\\hline
 FPC               & 96.83$\pm$0.86   & {\bf 99.91}$\pm$0.20  & {\bf 94.12}$\pm$0.39   & {\bf 99.27}$\pm$0.28    & 97.64$\pm$0.11    & 86.52$\pm$0.30    & 98.77$\pm$0.06    & 81.81$\pm$0.23    & 99.24$\pm$0.06   \\\hline
\end{tabular}
\end{center}
 \label{Tab:error}
\end{table*}

\begin{table*}
\caption{Comparison results between the proposed method and existing methods on UCI data sets in terms of AUC. }
\begin{center}
% \tiny
\begin{tabular}{|c|c|c|c|c|c|c|c|c|c|}\hline
%   \multirow{2}{*}
   {Data set}
%   &
%\multicolumn{5}{|c|}{TestAcc (std.) \%} \\
%\cline{2-6}
                   & breast           & banknote        & seismic          & musk2             & HTRU2             & MAGIC             & occupancy         & default           & Skin             \\\hline
 SVM-RBF           & 0.995$\pm$2e-3   & 0.998$\pm$2e-3  & 0.762$\pm$3e-2   & 0.942$\pm$1e-3    & 0.977$\pm$4e-3    & 0.882$\pm$2e-3    & 0.992$\pm$2e-4    & 0.752$\pm$3e-3    & 0.990$\pm$2e-4   \\\hline
 SVM-Poly          & 0.993$\pm$2e-3   & 0.995$\pm$2e-3  & 0.755$\pm$4e-2   & 0.956$\pm$1e-3    & 0.975$\pm$5e-3    & 0.908$\pm$2e-3    & 0.994$\pm$3e-4    & 0.758$\pm$3e-3    & 0.992$\pm$8e-5   \\\hline
 linear-SVM        & 0.986$\pm$1e-3   & 0.989$\pm$3e-3  & 0.752$\pm$3e-2   & 0.933$\pm$2e-3    & 0.968$\pm$4e-3    & 0.880$\pm$2e-3    & 0.993$\pm$2e-4    & 0.738$\pm$4e-3    & 0.988$\pm$5e-4   \\\hline
 $\ell_1$-SVM      & 0.987$\pm$2e-3   & 0.994$\pm$2e-3  & 0.758$\pm$3e-2   & 0.952$\pm$1e-3    & 0.972$\pm$5e-3    & 0.879$\pm$4e-3    & 0.989$\pm$4e-4    & 0.751$\pm$3e-3    & 0.989$\pm$4e-4   \\\hline
 RF                & 0.994$\pm$2e-3   & 0.998$\pm$1e-3  & 0.753$\pm$3e-2   & 0.968$\pm$1e-3    & 0.977$\pm$2e-3    & 0.926$\pm$2e-3    & {\bf 0.998}$\pm$2e-4    & 0.732$\pm$5e-3    & {\bf 1.000}$\pm$0      \\\hline
 PROX-PFGM          & 0.995$\pm$8e-4   & 0.998$\pm$1e-3  & 0.771$\pm$2e-2   & 0.961$\pm$1e-3    & {\bf 0.978}$\pm$4e-3    & 0.921$\pm$2e-3    & 0.996$\pm$2e-4    & {\bf 0.758}$\pm$2e-3    & 0.998$\pm$1e-4   \\\hline
 FALKON            & 0.995$\pm$2e-3   & 0.998$\pm$1e-3  & 0.768$\pm$2e-2   & 0.973$\pm$1e-3    & 0.976$\pm$3e-3    & {\bf 0.927}$\pm$2e-3    & 0.996$\pm$2e-4    & 0.752$\pm$4e-3    & 0.993$\pm$5e-4   \\\hline
 RR-Fourier        & 0.992$\pm$3e-3   & 0.996$\pm$2e-3  & 0.761$\pm$3e-2   & 0.956$\pm$2e-3    & 0.971$\pm$4e-3    & 0.913$\pm$3e-3    & 0.992$\pm$2e-4    & 0.746$\pm$4e-3    & 0.991$\pm$4e-4   \\\hline
 TS                & 0.988$\pm$3e-3   & 0.992$\pm$3e-3  & 0.758$\pm$4e-2   & 0.948$\pm$1e-3    & 0.971$\pm$3e-3    & 0.915$\pm$2e-3    & 0.990$\pm$3e-4    & 0.741$\pm$3e-3    & 0.989$\pm$1e-4   \\\hline
 FPC               & {\bf 0.996}$\pm$2e-3   & {\bf 1.000}$\pm$0     & {\bf 0.780}$\pm$3e-2   & {\bf 0.998}$\pm$7e-4    & 0.976$\pm$4e-3    & 0.916$\pm$2e-3    & 0.995$\pm$3e-4    & 0.744$\pm$3e-3    & 0.999$\pm$4e-5   \\\hline
\end{tabular}
\end{center}
 \label{Tab:auc}
\end{table*}

\begin{table*}
\caption{Training time (in seconds) of different algorithms on UCI data sets. }
\begin{center}
% \scriptsize
\begin{tabular}{|c|c|c|c|c|c|c|c|c|c|}\hline
%   \multirow{2}{*}
Data set           & breast    & banknote  & seismic & musk2   & HTRU2   & MAGIC    & occupancy   & default   & Skin             \\\hline
 SVM-RBF           &5.42        &8.35      &30.12   &1350.4    &638.8    &4561.8    &846.0     &9817.2     &9760.2    \\\hline
 SVM-Poly          &1.10        &6.07      &12.94   &845.89    &78.55    &12011.5   &2991.6    &35902.2    &851.6   \\\hline
 linear-SVM        &0.52        &2.32      &6.12    &303.12    &40.15    &2018.5    &362.1     &6754.2     &246.7   \\\hline
 $\ell_1$-SVM      &0.64        &2.28      &4.55    &315.24    &50.15    &819.23    &402.15    &1892.3     &252.6   \\\hline
 RF                &0.99        &1.21      &1.86    &7.36      &8.11     &18.18     &6.57      &30.65      &64.23      \\\hline
 PROX-PFGM         &0.38        &2.23      &4.15    &102.5     &36.1     &129.3     &96.5      &387.1      &114.5   \\\hline
 FALKON            &0.068       &0.37     &1.23    &2.84      &1.43      &3.21      &2.34      &2.21       &15.63   \\\hline
 RR-Fourier        &0.08        &0.12     &1.08    &3.64      &4.21      &3.89      &4.06      &3.89       &18.64   \\\hline
 TS                &0.018        &0.15     &0.62    &1.23      &1.65      &2.03      &1.46      &2.32       &10.13   \\\hline
 FPC               &{\bf 3e-6}        &{\bf 7.5e-3}   &{\bf 0.025}   &{\bf 2.91}      &{\bf 0.34}      &{\bf 0.58}      &{\bf 0.075}     &{\bf 0.32}       &{\bf 7.74}   \\\hline
\end{tabular}
\end{center}
 \label{Tab:training-time}
\end{table*}

\section{Real data experiments}
\label{sc:realdata}

In this section, we show the effectiveness of the proposed FPC via a series of experiments on nine UCI data sets covering various areas with medium sizes, and two massive data sets arising from the exotic particle discovery application in high-energy physics, as well as
three high-dimensional data sets from the application of image classification.
%a real application, i.e., image classification.

\subsection{UCI data sets}
\label{sc:UCIdata}

\subsubsection{Experimental setting}
In the following, we describe the setting of the experiments.

{\bf Samples}: All data sets are from:
\url{https://archive.ics.uci.edu/ml/datasets.html}.
The sizes of  data sets are listed in Table \ref{Tab:uci-data}.
For each data set, we used $50\%$, $25\%$ and $25\%$ samples as the training, validation and testing sets, respectively.

{\bf Competitors:}
We evaluated the effectiveness of FPC via comparing with nine baselines including
support vector machine (SVM) method with a radial basis function (\textit{SVM-RBF}), SVM with a polynomial (\textit{SVM-Poly}) kernel, linear-SVM \cite{Chang2010-linearsvm} with a linear kernel, $\ell_1$-SVM \cite{Zhu2003}, random forest (\textit{RF}) \cite{Breiman2001}, feature generating machine using polynomial feature mappings (\textit{PROX-PFGM}) \cite{Tan2014-FGM}, an efficient Nystr\"{o}m method, i.e., \textit{FALKON} \cite{Rudi2017}, and two random feature methods, i.e., the method suggested in \cite{Rahimi2007} with random Fourier features denoted as \textit{RR-Fourier},  and tensor sketching method with the kernel $(1+x\cdot x')^2$ denoted as \textit{TS} \cite{Pham2013}.
We used the \textit{libsvm} toolbox\footnote{\url{https://www.csie.ntu.edu.tw/~cjlin/libsvm/}} to implement SVM-RBF and SVM-Poly, and the \textit{LIBLinear} toolbox\footnote{\url{http://www.csie.ntu.edu.tw/cjlin/liblinear/}} to implement linear-SVM, $\ell_1$-SVM and TS.
%from the website: %\textit{https://www.csie.ntu.edu.tw/~cjlin/libsvm/}.

{\bf Implementation:}
For FPC, we set $\alpha =1$, $\beta=1$ and the initialization $p^0=(0,y,0)$ in the ADMM algorithm;
the stopping criterion of ADMM in FPC was set as the maximal iterations less than $5$,
which was generally adequate as shown in the previous simulations;
as shown by Theorem \ref{Thm:main}, the polynomial degree $s$ is empirically selected from the range $(1, s_{\max})$, where
$s_{\max}:= \min\left\{[m^{1/d}], 10\right\}$;
%$s_{\max}:= \min\left\{\left\lceil (\frac{m}{\log m})^{1/d}\right\rceil, 10\right\}$;
for a pre-determined $s$, the number of centers $n$ was set as $n= \min\left\{\left(^{s+d}_{\ s}\right),m\right\}$;
%given the number of centers $n$,
the centers $\{\eta_i\}_{i=1}^n$ were selected according to \textit{Scheme 1}, i.e., the uniformly random way.
These default settings of FPC in UCI data sets were determined by the previous simulations as well as our sensitivity studies on them over these nine UCI data sets later as presented in {\it Appendix F}.

For both {\textit SVM-RBF} and {\textit SVM-Poly}, the ranges of parameters $(c,g)$ involved in \textit{libsvm} were determined via a grid search on the region $[2^{-5}, 2^5] \times [2^{-5}, 2^5]$ in the logarithmic scale,
while for {\textit SVM-Poly}, the kernel parameter $s$ was selected from the interval [1, 10] via a grid search with 10 candidates, i.e., $\{1, 2, \ldots, 10\}$.
We used the default stopping criteria of \textit{libsvm} for these methods.
For linear-SVM and $\ell_1$-SVM, we used the default settings of \textit{LIBLinear}.
For \textit{RF}, the number of trees used was determined from the interval $[2, 20]$ via a grid search with 10 candidates, i.e., $\{2, 4, \ldots, 20\}$.
For PROX-PFGM, we used the same settings as that in \cite{Tan2014-FGM}.
For FALKON, RR-Fourier and TS, we used the same settings as that in the associated literature \cite{Rudi2017,Rahimi2007,Pham2013}, where the number of features was selected from the interval $[200, 2000]$ via a grid search with 10 candidates, i.e., $\{200, 400, \ldots, 2000\}$.

For each data set, we ran $20$ times for all algorithms, and then recorded their averages and standard deviations (std) of testing accuracies (\textit{Accuracy}) \footnote{Testing accuracy is defined as the percentage of the correct classification.} and area under curve (AUC)\footnote{AUC is defined as the area under the receiver operating characteristic (ROC) curve and is implemented by the Matlab function ``perfcurve'' with default parameters in this paper.}, as well as the averages of training time (\textit{Time}).
%and the \textit{sparsity level} \footnote{For SVMs, the sparsity level is the number of the support vectors, for FPC, the sparsity level is the number of selected centers, and for random forest, the sparsity level is the number of selected trees.}.
Here, we used AUC as an important evaluation metric since some UCI data sets are imbalanced as shown in Table \ref{Tab:uci-data}.

\subsubsection{Experimental results}

\begin{figure}[!t]
\begin{minipage}[b]{0.49\linewidth}
\centering
\includegraphics*[scale=0.32]{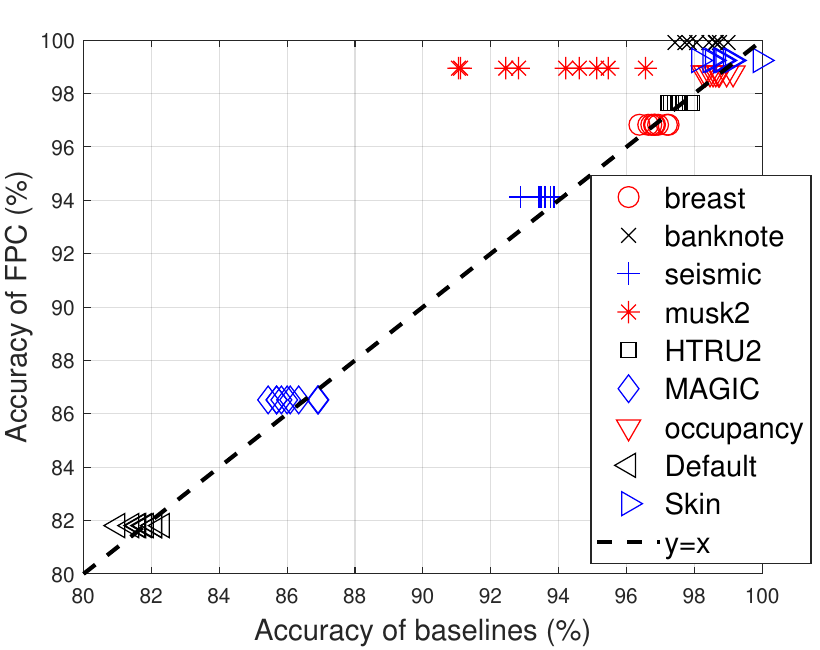}
%  \vspace{-.5cm}
\centerline{{\small (a) Test accuracy}}
\end{minipage}
\hfill
\begin{minipage}[b]{0.49\linewidth}
\centering
\includegraphics*[scale=0.32]{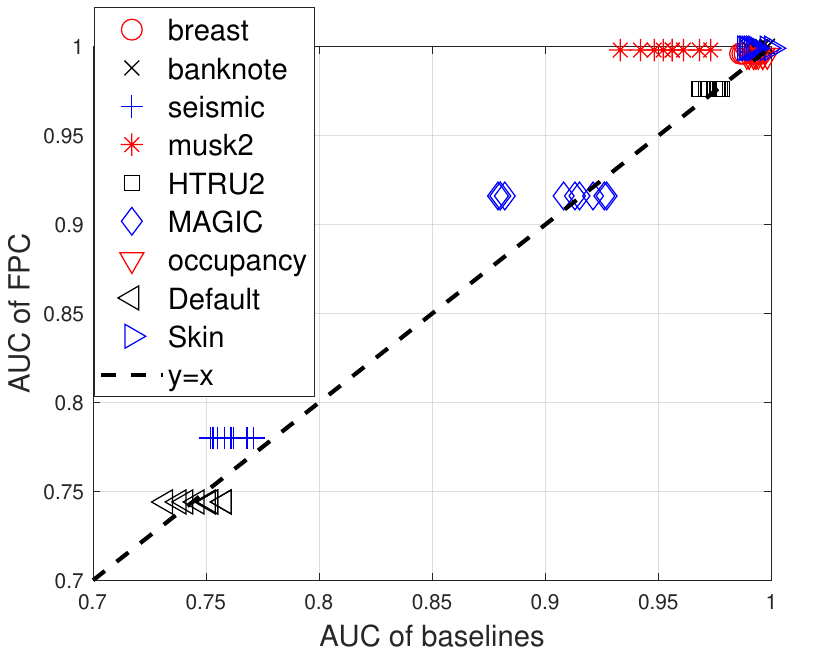}
%  \vspace{-.5cm}
\centerline{{\small (b) AUC}}
\end{minipage}
\hfill
\caption{ Performance of FPC versus those of nine competitors in terms of both testing accuracy and AUC on UCI data sets.
 }
\label{fig:uci-comparison}
\end{figure}

The experimental results of UCI data are reported in Tables \ref{Tab:error}--\ref{Tab:training-time}, which represent the testing accuracy, AUC and training time of all these methods, respectively.
To demonstrate more clearly the effectiveness of the proposed FPC with these baselines, the testing accuracy and AUC of the proposed method versus those of the nine baselines on the nine UCI data sets are respectively shown graphically in Fig. \ref{fig:uci-comparison}(a) and (b), where the x- and y-axes, respectively represent the test accuracy (AUC) of the baselines and that of FPC, and different colors represent the results of different data sets. The dashed line with the equation $y=x$ is a baseline to give an intuitive comparison between the competitors and the proposed method.

In terms of test accuracy, we can observe from Table \ref{Tab:error} and Fig. \ref{fig:uci-comparison}(a) that the proposed method achieves the best performance over three data sets, i.e., \textit{banknote, musk2, seismic}, while is comparable to the competitors over the rest six data sets.
Particularly, as shown in  Table \ref{Tab:error}, the testing accuracy of FPC is significantly better than those of competitors over the \textit{musk2} data set, with about 3.7\% improvement in terms of testing accuracy when compared to the best result yielded by the competitors, that is, 95.56\% (RF) vs. {\bf 99.27}\% (FPC).

In terms of AUC, it can be observed from Table \ref{Tab:auc} and Fig. \ref{fig:uci-comparison}(b) that the proposed FPC achieves the best performance over four data sets, i.e.,  \textit{breast, banknote, musk2, seismic}, while is comparable to the baselines over the rest five data sets. Moreover, except \textit{musk2} and \textit{default} data sets, the proposed FPC achieves relatively very high AUC (greater than 0.916) for all the rest UCI data sets, though some data sets are very imbalanced (say, the percentages of majority class of \textit{musk2, HTRU2, occupancy} and \textit{Skin} data sets are 84.59\%, 90.84\%, 76.90\%, and 79.25\%, respectively). This demonstrates the effectiveness of the proposed FPC.

As far as the training time is concerned, we can observe from Table \ref{Tab:training-time} that the training time of FPC is much less than that of the classical kernel approaches, especially when the size of data exceeds ten thousands, and the proposed FPC is also faster than some existing Nystr$\ddot{o}$m and random feature methods including FALKON, RR-Fourier and TS.

\begin{table}
\caption{Details of two massive data sets, i.e., SUSY and HIGGS, and three high-dimensional data sets, i.e., Gisette, MNIST38 and Dogs\_vs\_Cats, where $n_{\mathrm{train}}$, $n_{\mathrm{valid}}$ and $n_{\mathrm{test}}$ represent the number of samples for training, validation and testing, respectively.
 $\dag$ For SUSY, we only consider eight low-level features.}
\begin{center}
\scriptsize
\begin{tabular}{|l|c|c|c|c|}\hline
  Data sets      & $n_{\mathrm{train}}$  & $n_{\mathrm{valid}}$  & $n_{\mathrm{test}}$  & \#Attributes\\\hline
 SUSY$\dag$      &4,000,000     & 500,000     & 500,000      &8\\ \hline
 HIGGS           &10,000,000    & 500,000     & 500,000     &21\\ \hline
 Gisette         &6,000         &1,000    &6,500     &5,000\\ \hline
 MNIST38         &35,000        &5,000    &22,581    &784\\ \hline
 Dogs\_vs\_Cats  &2,000         &1,911    &25,000    &1,024\\ \hline
\end{tabular}
\end{center}
 \label{Tab:other-data}
\end{table}

\subsection{Massive data sets: SUSY and HIGGS}
\label{sc:susy-higgs}

The field of high-energy physics is devoted to the study of  elementary constituents of matter.
By investigating the structure of matter and the laws that govern its interactions,
this field strives to discover  fundamental properties of the physical universe.
The primary tools of experimental high-energy physicists are modern accelerators,
which collide protons and/or antiprotons to create exotic particles that occur only at extremely high-energy densities.
Observing these particles and measuring their properties may yield critical insights about the very nature of matter.
Finding these rare particles requires solving difficult signal-versus-background classification problems.
In the following, we considered two benchmark massive data sets in this field, i.e., supersymmetry particles (SUSY) and Higgs bosons (HIGGS), as studied in \cite{Baldi2014}.
These two data sets are available from the links: \textit{https://archive.ics.uci.edu/ml/datasets/SUSY}, and   \textit{https://archive.ics.uci.edu/ml/datasets/HIGGS}, respectively. Detailed information of these two data sets can be found in Table \ref{Tab:other-data}.

For both data sets, the parameter settings of FPC were the same as those in Section \ref{sc:UCIdata}, except $n$ was tuned from some ranges via a grid search (the ranges in SUSY and Higgs are $[1500, 1700]$ and $[300, 500]$, respectively, with 11 candidates). Since the training processes of those classical SVM methods and random forest involved in UCI data sets are very time-consuming for these two massive data sets, we only implemented FPC and compared the performance of FPC with the other four faster methods including PROX-PFGM \cite{Tan2014-FGM}, FALKON \cite{Rudi2017}, RR-Fourier \cite{Rahimi2007} and TS \cite{Pham2013}, whose parameter settings were the same as those in Section \ref{sc:UCIdata}. The testing accuracy, AUC and training time in minutes are presented in Table \ref{Tab:massive-data-comp}.

\begin{table}
\caption{Performance comparison of different algorithms on two massive data sets, i.e., SUSY and HIGGS, where testing accuracy and training time are recorded in percentages and minutes, respectively. The best and second best results are marked in bold and underline, respectively.}
%\vspace{0.1cm}
%\tiny
\scriptsize
\begin{center}
\begin{tabular}{|c|c|c|c|c|c|c|}\hline
dataset &\multicolumn{3}{|c|}{SUSY} &\multicolumn{3}{|c|}{HIGGS}\\\hline
metric       &Accuracy  & AUC     & Time(m)       &Accuracy   &AUC       & Time(m)  \\\hline
Prox-PFGM     &78.88         & 0.859   &62.4          &64.87          &0.826     &212.9\\\hline
FALKON       &{\bf 79.12}         & $\underline{0.871}$   &46.7              &{\bf 65.42}          &{\bf 0.883}   &163.4\\\hline
RR-Fourier   &78.34         & 0.847   &16.5           &64.65          &0.814     &21.4\\\hline
TS           &77.55         & 0.828   &$\underline{11.6}$           &64.17          &0.814     &$\underline{18.3}$\\\hline
FPC          &$\underline{78.99}$         & {\bf 0.876}   & {\bf 10.2}          &$\underline{65.39}$          &$\underline{0.878}$     &{\bf 4.6}\\\hline
\end{tabular}
\end{center}
\label{Tab:massive-data-comp}
%\vspace{-.8cm}
\end{table}

From Table \ref{Tab:massive-data-comp}, in terms of testing accuracy,
FPC outperforms the other concerned methods except FALKON, while FPC is comparable to FALKON in terms of AUC and achieves the best AUC result over SUSY.
In the perspective of training time, FPC is faster than the concerned four competitors over these two massive data sets.
Specifically, when compared to FALKON, a typical Nystr$\ddot{o}$m method, the proposed FPC is much faster but with the comparable generalization performance,
while when compared to those two typical random feature methods, i.e., RR-Fourier and TS, the proposed FPC outperforms them in terms of all three evaluation metrics.
These demonstrate the effectiveness of the proposed FPC.

\subsection{High-dimensional data sets}

%\subsection{Image classification on Dogs vs. Cats competition data}

\begin{table}
\caption{Performance comparison of different algorithms on three high-dimensional data sets, i.e., Gisette, MNIST38 and Dogs\_vs\_Cat, where the testing accuracy and training time are recorded in percentages and seconds respectively.}
%\vspace{0.1cm}
%\tiny
\scriptsize
\begin{center}
\begin{tabular}{|c|c|c|c|c|c|c|}\hline
dataset &\multicolumn{2}{|c|}{Gisette}     &\multicolumn{2}{|c|}{MNIST38}  &\multicolumn{2}{|c|}{Dogs\_vs\_Cats}\\\hline
metric       &Accuracy     & Time   &Accuracy    & Time     &Accuracy     & Time \\\hline
Prox-PFGM    &96.62            & 28.4      & 99.1            &32.6         &96.32            & 14.3 \\\hline
FALKON       &{\bf 96.66}            & 32.6      & {\bf 99.3}            &21.5         &$\underline{96.73}$            & 8.7\\\hline
RR-Fourier   &96.52            & 17.3      & 99.2            &15.6         &96.68            & 6.5\\\hline
TS           &96.44            & $\underline{11.6}$      & 98.9            &$\underline{8.6}$          &96.12            & $\underline{4.3}$\\\hline
FPC          &$\underline{96.63}$            & {\bf 10.2}      & $\underline{99.2}$            &{\bf 6.2}          &{\bf 96.82}            & {\bf 1.08}\\\hline
\end{tabular}
\end{center}
\label{Tab:highdim-data-comp}
%\vspace{-.8cm}
\end{table}

In this section, we show the effectiveness of the proposed FPC over three high-dimensional data sets, which are all from the application of image classification.

The first one is the \textit{Gisette} data set, which is available from \url{https://archive.ics.uci.edu/ml/datasets/Gisette}. Gisette is constructed for a handwritten digit recognition problem, which aims to separate the highly confusible digits `4' and `9'. As presented in Table \ref{Tab:other-data}, the feature dimension of Gisette is the very high dimension 5,000.
The second one is the \textit{MNIST38} data set, which consists of the digital images of 3 and 8 from the MNIST data set\footnote{The data set is available from \url{https://www.csie.ntu.edu.tw/~cjlin/libsvmtools/datasets/}.}. The feature dimension of the MNIST38 data set is 782.
The third one is the \textit{Dogs\_vs\_Cats} data set\footnote{https://www.kaggle.com/c/dogs-vs-cats.}, which aims to classify whether images contain either a dog or a cat. Different from Gisette and MNIST38, features of the \textit{Dogs\_vs\_Cats} data set are yielded from pre-trained GoogLeNet \cite{Szegedy2015} with 1024 dimensions (corresponding to $32\times 32$ image patches), as similarly done in the literature \cite{Rudi2017}.
Detailed information of these data sets is presented in Table \ref{Tab:other-data}.

The parameter settings of FPC are the same as those in massive data sets except that the number of selected centers $n$ is determined from the interval $[200, 2000]$ via a grid search way with 10 candidates as the same to those for the concerned four competitors, i.e., PROX-PFGM, FALKON, RR-Fourier and TS. Since the considered three high-dimensional data sets are not imbalanced, we only compared the performance of FPC with competitors in terms of both testing accuracy in percentages and training time in seconds. The comparison results over these three high-dimensional data sets can be found in Table \ref{Tab:highdim-data-comp}.

From Table \ref{Tab:highdim-data-comp}, we can observe that the proposed FPC achieves the best result over the Dogs\_vs\_Cats data set and the second best results over the other two data sets in terms of testing accuracy, while in the perspective of training time, FPC is faster than the concerned competitors over all three data sets.  In particular, when compared to these two typical random feature methods, i.e., RR-Fourier \cite{Rahimi2007} and TS \cite{Pham2013}, the proposed FPC outperforms them in terms of both testing accuracy and training time. These demonstrate the effectiveness of the proposed method.

\section{Conclusion}
\label{sc:conclusion}

The design of efficient classification methods for massive data classification is an important topic in the era of big data.
Classical kernel approaches are not scalable enough to handle   massive data, mainly because they are required to map the original data into a very high dimensional space whose dimension is the size of samples, which is commonly ``unnecessarily high'' for classification.
Due to such a kernel trick, the computational burden and storage of the classical kernel approaches are generally unaffordable when dealing with the massive data classification problem.
In this paper, we propose a fast, efficient learning scheme called \textit{FPC} for massive data classification based on  polynomial kernels.
We exploit some subsampling scheme, based on which an effective feature mapping can be constructed from  polynomial kernels.
Instead of mapping the original data into the very high dimensional space, the constructed feature mapping generally maps the original data into a relatively low dimensional space.
Then we find   classifiers on such lower dimensional feature spaces efficiently via exploiting the alternating direction method of multipliers (ADMM).
The effectiveness of the proposed learning scheme is verified in both theory and numerical experiments.
Theoretically, we justify that the suggested learning scheme preserves almost the same generalization power of SVM.
Numerically, the proposed FPC is much faster than  SVM  but does not sacrifice its generalization.
This shows that FPC brings some extent of possibility for handling massive data classification problems.

\section*{Acknowledgments}
The authors would like to thank Drs. Yao Wang and Xu Liao for sharing the \textit{Dogs\_vs\_Cats} competition data set. The work of Jinshan Zeng was partially supported by the National Natural Science Foundation of China [Grant No. 61977038] and by the Thousand Talents Plan of Jiangxi Province [Grants No. jxsq2019201124]. The work of Shao-Bo Lin was supported by the National Natural Science Foundation of China [Grant No. 61876133], and The work of Ding-Xuan Zhou is supported partially by NSFC/RGC Joint Research Scheme [RGC Project No. N\_Cityu 102/20 and NSFC Project No. 12061160462], Laboratory for AI-Powered Financial Technologies and by the Hong Kong Institute for Data Science.

\appendices
\section*{Appendix A. Proof of Theorem \ref{Thm:conv-admm}}
\label{Appendix:Proof-Conv-ADMM}

According to \cite{Facchinei-VI2003-si}, it is easy to derive the variational inequality (VI) reformulation of \eqref{Eq:opt_prob_admm} as follows:
Find $p^* = (u^*,v^*,w^*) \in \Omega:=\mathbb{R}^n \times \mathbb{R}^m \times \mathbb{R}^m$ such that for any $p \in \Omega$,
\begin{align}
\label{Eq:VI}
\mathrm{VI}(\Omega,F,f):=f(v) - f(v^*) + (p-p^*)^T F(p^*) \geq 0,
\end{align}
where
\begin{align}
p = \left( \begin{array}{c}
u\\
v\\
w
\end{array}
\right),
\
F(p)=\left( \begin{array}{c}
A^Tw\\
-w\\
v-Au
\end{array}
\right). \label{Eq:def-p-FP}
\end{align}
Note that the mapping $F(p)$ is monotone because it is affine with a skew-symmetric matrix.
We denote by $\Omega^*$ the solution set of $\mathrm{VI}(\Omega,F,f)$.
By \cite{Facchinei-VI2003-si}, the VI formulation \eqref{Eq:VI} is equivalent to the constrained formulation \eqref{Eq:opt_prob_admm} in the sense that for any $p^* \in \Omega^*$,  $(u^*,v^*)$ is a minimizer of \eqref{Eq:opt_prob_admm} and vice versa, that is, if $(u^*,v^*)$ is a minimizer of \eqref{Eq:opt_prob_admm}, then there exists a $w^*$ such that $p^*=(u^*,v^*,w^*) \in \Omega^*$.
Furthermore, note from the relation between the problems \eqref{Eq:opt_prob} and \eqref{Eq:opt_prob_admm},
it is obvious that $\mathrm{VI}(\Omega,F,f)$ is equivalent to the original unconstrained problem \eqref{Eq:opt_prob}.

In order to prove Theorem \ref{Thm:conv-admm}, we need the following lemmas.

\begin{lemma}
\label{Lemm:lemma1-conv}
Let $\{p^k\}$ be the sequence generated by (\ref{Eq:update-u}), (\ref{Eq:update-v}), (\ref{Eq:update-w}). Then for any $k\in\mathbb N$ and $p\in \mathbb{R}^{n+2m}$, there holds
\begin{align}
\label{Eq:Lemma1-conv}
&f(v)-f(v^{k+1})+ \\
&(p-p^{k+1})^T\left[F(p^{k+1})+ \eta(v^k,v^{k+1}) + H(p^{k+1}-p^k) \right] \geq 0, \nonumber
\end{align}
where $H$ is defined in \eqref{Eq:def-H}, and
\begin{align}
\label{Eq:def-eta}
\eta(v^k,v^{k+1}):= \beta
\left(\begin{array}{c}
-A^T\\
{\bf I}_m\\
0
\end{array}
\right)
(v^k - v^{k+1}).
\end{align}
\end{lemma}

\begin{proof}
By the $u^{k+1}$-update \eqref{Eq:update-u}, the optimality of $u^{k+1}$ implies
\begin{align}
\label{Eq:opt-uk}
0
&= A^T\left[\beta(Au^{k+1}-v^k) + w^k\right] + \alpha (u^{k+1} - u^k) \nonumber\\
&= A^Tw^{k+1} - \beta A^T(v^k - v^{k+1}) + \alpha (u^{k+1}-u^k).
\end{align}
By the $v^{k+1}$-update \eqref{Eq:update-v}, the optimality of $v^{k+1}$ implies
\begin{align*}
0 &\in \partial f(v^{k+1}) - \left[\beta(Au^{k+1}-v^{k+1})+w^k\right] \\
& = \partial f(v^{k+1})-w^{k+1}.
\end{align*}
By the convexity of $f$, the above equality shows for any $k\in \mathbb{N}$,
\begin{align}
\label{Eq:opt-vk}
f(v) - f(v^{k+1}) - (v-v^{k+1})^Tw^{k+1} \geq 0, \ \forall v\in \mathbb{R}^n.
\end{align}
It follows from \eqref{Eq:update-w} that
\begin{align}
\label{Eq:opt-wk}
\beta^{-1}(w^{k+1}-w^k) - (Au^{k+1}-v^{k+1})=0.
\end{align}
Combining \eqref{Eq:opt-uk}, \eqref{Eq:opt-vk} and \eqref{Eq:opt-wk} together, for any $p$, we have
\begin{align*}
&f(v)-f(v^{k+1})
+
\left(
\begin{array}{c}
u-u^{k+1}\\
v-v^{k+1}\\
w-w^{k+1}
\end{array}
\right)^T \\
&
\Bigg\{
\left(
\begin{array}{c}
A^Tw^{k+1}-\beta A^T(v^k-v^{k+1})\\
-w^{k+1}\\
v^{k+1}-Au^{k+1}
\end{array}
\right)\\
% \right.
&+
% \left.
\left(
\begin{array}{c}
\alpha (u^{k+1}-u^k)\\
0\\
\beta^{-1}(w^{k+1}-w^k)
\end{array}
\right)
\Bigg\}
%\right\}
\geq 0,
\end{align*}
which can be rewritten as
\begin{align}
&f(v)-f(v^{k+1}) \label{Eq:Lemma1-conv1}
+
\left(
\begin{array}{c}
u-u^{k+1}\\
v-v^{k+1}\\
w-w^{k+1}
\end{array}
\right)^T
% \left\{
\Bigg\{
\left(
\begin{array}{c}
A^Tw^{k+1}\\
-w^{k+1}\\
v^{k+1}-Au^{k+1}
\end{array}
\right)\\
&+\beta
\left(
\begin{array}{c}
-A^T(v^k-v^{k+1})\\
v^k-v^{k+1}\\
0
\end{array}
\right)
+
\left(
\begin{array}{c}
\alpha (u^{k+1}-u^k)\\
\beta(v^{k+1}-v^k)\\
\beta^{-1}(w^{k+1}-w^k)
\end{array}
\right)
\Bigg\}
% \right\}
\geq 0.\nonumber
\end{align}
By the notations of $F(p)$, $\eta(v^k,v^{k+1})$ and $H$, we get \eqref{Eq:Lemma1-conv} immediately.
\end{proof}

\begin{lemma}
\label{Lemm:strict-contract}
Let $\{p^k\}$ be the  sequence generated by (\ref{Eq:update-u}), (\ref{Eq:update-v}), (\ref{Eq:update-w}), then for any $k\in \mathbb{N}$, and $  p^* \in \Omega^*$, there holds
\begin{align}
\label{Eq:strict-contract}
\|p^{k+1}-p^*\|_H^2 \leq \|p^k - p^*\|_H^2 - \|p^k - p^{k+1}\|_H^2.
\end{align}
\end{lemma}

\begin{proof}
Since $p^* \in \Omega^*$, it follows from \eqref{Eq:VI} that
\begin{align*}
f(v^{k+1}) - f(v^*) + (p^{k+1}-p^*)^TF(p^*) \geq 0.
\end{align*}
By the monotonicity of $F$, we have
\begin{align}
\label{Eq:nonnegative-part1}
f(v^{k+1}) - f(v^*) + (p^{k+1}-p^*)^TF(p^{k+1}) \geq 0.
\end{align}
Note that \eqref{Eq:opt-vk} is satisfied for both $k$ and $k+1$, that is, for any $v\in \mathbb{R}^n$,
\begin{align*}
& f(v) - f(v^{k+1}) - (v-v^{k+1})^Tw^{k+1} \geq 0,\\
& f(v) - f(v^{k}) - (v-v^{k})^Tw^{k} \geq 0.
\end{align*}
Setting $v=v^k$ and $v = v^{k+1}$ in the first and second inequalities, respectively, and then adding them yields
\begin{align}
\label{Eq:innerproduc-v-w}
\langle v^k - v^{k+1}, w^k - w^{k+1}\rangle \geq 0.
\end{align}
By using the notation of $\eta(v^k,v^{k+1})$, the relation $v^* = Au^*$, and the $w^k$-update \eqref{Eq:update-w}, we have
\begin{align}
\label{Eq:nonnegative-part2}
&(p^{k+1}-p^*)^T\eta(v^k - v^{k+1}) \nonumber\\
&=\beta (v^k - v^{k+1})^T \left[ -A(u^{k+1}-u^*) + (v^{k+1}-v^*)\right] \nonumber\\
&= \beta (v^k - v^{k+1})^T (v^{k+1}-Au^{k+1}) \nonumber\\
&=(v^k - v^{k+1})^T (w^k - w^{k+1}) \geq 0,
\end{align}
where the final inequality follows from \eqref{Eq:innerproduc-v-w}.

Setting $p=p^*$ in \eqref{Eq:Lemma1-conv}, we have
\begin{align}
\label{Eq:nonnegative-crossterm}
&(p^{k+1}-p^*)^TH(p^k - p^{k+1}) \nonumber\\
&\geq f(v^{k+1})-f(v^*)+(p^{k+1}-p^*)^TF(p^{k+1}) \nonumber\\
&+(p^{k+1}-p^*)^T\eta(v^k,v^{k+1})\geq 0,
\end{align}
where the final inequality holds due to  \eqref{Eq:nonnegative-part1} and \eqref{Eq:nonnegative-part2}.

By \eqref{Eq:nonnegative-crossterm}, we have
\begin{align*}
&\|p^k - p^*\|_H^2 = \|(p^{k+1}-p^*)+(p^k - p^{k+1})\|_H^2\\
& = \|p^{k+1} - p^*\|_H^2 + \|p^k - p^{k+1}\|_H^2 \\
&+ 2(p^{k+1}-p^*)^TH(p^k - p^{k+1})\\
&\geq \|p^{k+1} - p^*\|_H^2+ \|p^k - p^{k+1}\|_H^2.
\end{align*}
This finishes the proof of this lemma.
\end{proof}

\begin{proof}{[Proof of Theorem \ref{Thm:conv-admm}]}
By Lemma \ref{Lemm:strict-contract}, the generated sequence $\{p^k\}$ is bounded. Actually, it is contained by the compact set
\[
\{p\in \mathbb{R}^{n+2m}: \|p-p^*\|_H^2 \leq \|p^0 - p^*\|_H^2\}
\]
where $p^*$ is an arbitrary point in $\Omega^*$, $p^0 \in \mathbb{R}^{n+2m}$ is the initialization.
Therefore, it is obvious that $\{p^k\}$ has at least one cluster point, say $p^{\infty}$,
and we assume that a subsequence $\{p^{k_j}\}_{j\in \mathbb{N}}$ converges to $p^{\infty}$.
Note that \eqref{Eq:strict-contract} directly implies $\|p^{k_j+1}-p^{k_j}\|_H^2 \rightarrow 0$ when $k_j \rightarrow \infty.$
Thus, taking the limit over $k_j \rightarrow \infty$ in \eqref{Eq:Lemma1-conv}, we have that $p^{\infty}$ is a solution of $\mathrm{VI}(\Omega,F,f)$ defined in \eqref{Eq:VI}, and thus, $u^{\infty}$ is a global minimizer of \eqref{Eq:opt_prob}.
Again by \eqref{Eq:strict-contract}, it implies that $p^{\infty}$ is the unique cluster point of the sequence $\{p^k\}$.
Thus, $\{p^k\}$ converges to $p^{\infty}$, a solution of \eqref{Eq:VI}, starting from any initial point $p^0$.
As a consequence, $\{u^k\}$ converges to a global minimizer of the original problem \eqref{Eq:opt_prob}.
This finishes the proof of this theorem.
\end{proof}

\section*{Appendix B. Proof of Theorem \ref{Thm:conv-rate-admm}}
\label{Appendix:Proof-ConvRate-ADMM}

For the convenience of analysis, we introduce some notations
\begin{align}
&M = \left(
\begin{array}{ccc}
{\bf I}_n &0 &0\\
0 &{\bf I}_m &0\\
0 &-\beta {\bf I}_m &{\bf I}_m
\end{array}
\right),
\label{Eq:def-M}
\\
&Q = \left(
\begin{array}{ccc}
\alpha{\bf I}_n &0 &0\\
0 &\beta{\bf I}_m &0\\
0 &-{\bf I}_m &\beta^{-1}{\bf I}_m
\end{array}
\right).
\label{Eq:def-Q}
\end{align}
From the definitions of $M$ and $Q$, it is easy to show that
\begin{align}
\label{Eq:relation-M-Q}
Q=HM,\ (Q^T+Q)-M^THM \succeq 0.
\end{align}
Moreover, we introduce an auxiliary variable $\tp^k$ defined as
\begin{align}
\label{Eq:def-tpk}
\tp^k =
\left(
\begin{array}{c}
\tu^k\\
\tv^k\\
\tw^k
\end{array}
\right)
=
\left(
\begin{array}{c}
u^{k+1}\\
v^{k+1}\\
w^k + \beta (Au^{k+1}-v^k)
\end{array}
\right),
\end{align}
then we have
\begin{align}
\label{Eq:tpk-pk}
p^{k+1} = p^k - M(p^k - \tp^k).
\end{align}

We still need the following lemma to show the monotonicity of the sequence $\{\|p^{k+1}-p^k\|_H^2\}$.

\begin{lemma}
\label{Lemm:monotone}
Let $\{p^k\}$ be the sequence generated by Algorithm \ref{alg1}, then for any $k\geq 1$, there holds
\begin{align}
\label{Eq:monotone}
\|p^{k+1}-p^k\|_H^2 \leq \|p^k - p^{k-1}\|_H^2.
\end{align}
\end{lemma}

\begin{proof}
By using the definition \eqref{Eq:def-tpk} of $\tp^k$, and the facts
\begin{align*}
&\beta^{-1}(w^k - \tw^k) = -(Au^{k+1}-v^k)\\
&= -(A\tu^k - \tv^k) - (\tv^k - v^k),
\end{align*}
 \eqref{Eq:Lemma1-conv1} can be rewritten as
\begin{align*}
&f(v)-f(\tv^k) + (p-\tp^k)^T
% \left\{
\Bigg\{
\left(
\begin{array}{c}
A^T\tw^k\\
-\tw^k\\
\tv^k-A\tu^k
\end{array}
\right)\\
% \right.
&+
% \left.
\left(
\begin{array}{c}
\alpha(\tu^k - u^k)\\
\beta(\tv^k-v^k)\\
-(\tv^k - v^k) + \beta^{-1}(\tw^k - w^k)
\end{array}
\right)
% \right\}
\Bigg\}
\geq 0, \ \forall p\in \mathbb{R}^{n+2m}.
\end{align*}
By the definition \eqref{Eq:def-Q}  of $Q$ and \eqref{Eq:def-p-FP} of $F(p)$, the above inequality yields for any $p\in \Omega$,
\begin{align}
\label{Eq:Lemma2-conv1}
&f(v)-f(\tv^k) + (p-\tp^k)^T\left[F(\tp^k) + Q(\tp^k-p^k) \right] \geq 0.
\end{align}
Note that \eqref{Eq:Lemma2-conv1} also holds when $k$ is replaced by $k+1$, and thus we have
\begin{align}
\label{Eq:Lemma2-conv1-k+1}
&f(v)-f(\tv^{k+1})
+ (p-\tp^{k+1})^T\left[F(\tp^{k+1}) + Q(\tp^{k+1}-p^{k+1}) \right] \nonumber\\
&\geq 0.
\end{align}
Setting $p=\tp^{k+1}$ and $p=\tp^{k}$ in \eqref{Eq:Lemma2-conv1} and \eqref{Eq:Lemma2-conv1-k+1}, respectively, we have
\begin{align*}
&f(\tv^{k+1})-f(\tv^k) + (\tp^{k+1}-\tp^k)^T\left[F(\tp^k) + Q(\tp^k-p^k) \right] \geq 0,
\end{align*}
and
\begin{align*}
&f(\tv^k)-f(\tv^{k+1})
+ (\tp^k-\tp^{k+1})^T\left[F(\tp^{k+1}) + Q(\tp^{k+1}-p^{k+1}) \right] \\
&\geq 0.
\end{align*}
Adding the above two inequalities and using the monotonicity of $F$ yields
\begin{align}
\label{Eq:Lemma2-conv2}
(\tp^k - \tp^{k+1})^T Q\left[(p^k - p^{k+1}) - (\tp^k - \tp^{k+1}) \right] \geq 0.
\end{align}
Adding the term
\[
\left[(p^k - p^{k+1}) - (\tp^k - \tp^{k+1}) \right]^TQ\left[(p^k - p^{k+1})-(\tp^k - \tp^{k+1}) \right]
\]
to both sides of \eqref{Eq:Lemma2-conv2} and using $p^TQp = \frac{1}{2}p^T(Q^T+Q)p$, we have
\begin{align*}
&(p^k - p^{k+1})^TQ\left[(p^k - p^{k+1})-(\tp^k - \tp^{k+1}) \right]\\
&\geq \frac{1}{2}\|(p^k - p^{k+1})-(\tp^k - \tp^{k+1})\|_{Q^T+Q}^2 \\
&=\frac{1}{2}\|(p^k - \tp^{k})-(\tp^{k+1} - \tp^{k+1})\|_{Q^T+Q}^2.
\end{align*}
By \eqref{Eq:relation-M-Q} and \eqref{Eq:tpk-pk}, the above inequality implies
\begin{align}
\label{Eq:Lemma2-conv3}
&(p^k - \tp^k)^TM^THM\left[(p^k - \tp^k)-(p^{k+1} - \tp^{k+1}) \right]\\
&\geq\frac{1}{2}\|(p^k - \tp^{k})-(\tp^{k+1} - \tp^{k+1})\|_{Q^T+Q}^2. \nonumber
\end{align}

Setting $a = M(p^k -\tp^k)$ and $b=M(p^{k+1}-\tp^{k+1})$ in the identity
\[
\|a\|_H^2 - \|b\|_H^2 = 2a^TH(a-b)-\|a-b\|_H^2,
\]
we obtain
\begin{align*}
&\|M(p^k-\tp^k)\|_H^2 - \|M(p^{k+1}-\tp^{k+1})\|_H^2\\
&=2(p^k -\tp^k)^TM^THM\left[(p^k-\tp^k)-(p^{k+1}-\tp^{k+1}) \right]\\
&-\|M\left[(p^k -\tp^k)-(p^{k+1}-\tp^{k+1}) \right]\|_H^2.
\end{align*}
Plugging \eqref{Eq:Lemma2-conv3} into the above inequality yields
\begin{align*}
&\|M(p^k-\tp^k)\|_H^2 - \|M(p^{k+1}-\tp^{k+1})\|_H^2\\
&\geq \|(p^k-\tp^k)-(p^{k+1}-\tp^{k+1})\|_{(Q^T+Q)}^2 \\
&- \|M\left[(p^k -\tp^k)-(p^k - \tp^{k+1}) \right]\|_H^2\\
&=\|(p^k - \tp^k)-(p^{k+1}-\tp^{k+1})\|_{\left[(Q^T+Q)-M^THM\right]}^2
 \geq 0,
\end{align*}
where the last inequality is due to $(Q^T+Q)-M^THM \succeq 0$ via \eqref{Eq:relation-M-Q}.
Furthermore, by \eqref{Eq:tpk-pk}, the above inequality implies \eqref{Eq:monotone}.
This finishes the proof of this lemma.
\end{proof}

Based on Lemmas \ref{Lemm:strict-contract} and \ref{Lemm:monotone}, we can prove Theorem \ref{Thm:conv-rate-admm} as follows.

\begin{proof}{[Proof of Theorem \ref{Thm:conv-rate-admm}]}
By Lemma \ref{Lemm:strict-contract}, we have
\begin{align}
\label{Eq:finite-sum}
\sum_{t=0}^{\infty} \|p^t - p^{t+1}\|_H^2 \leq \|p^0 - p^*\|_H^2, \ \forall p^* \in \Omega^*.
\end{align}
According to Lemma \ref{Lemm:monotone}, the sequence $\{\|p^t-p^{t+1}\|_H^2\}$ is non-increasing.
Thus, by \cite[Lemma 1.1]{Deng2017-si}, we can get the $o(1/k)$ convergence rate of $\|p^t-p^{t+1}\|_H^2$.
This finishes the proof.
\end{proof}

\section*{Appendix C. Proof of Theorem \ref{Thm:main}}
\label{Appendix:Proof-Learning-Theory}
Denote $\phi(t):=(1-t)_+$ and $\mathcal
E(f):=\int_Z\phi(yf(x))d\rho$ for   $f \in L_{\rho_X}^2$.
It can be found in \cite{Zhang2004-si}  that
\begin{equation}\label{comparison teorem2}
          \mathcal R(\mbox{sgn}(f))-\mathcal R(f_c)\leq  \mathcal E(f)-\mathcal
        E(f_\rho),
\end{equation}
where $f_\rho$ is defined by (\ref{definition of frho}). Thus, to prove Theorem \ref{Thm:main}, it suffices to bound  $\mathcal E(\pi f_{D,n})-\mathcal
        E(f_\rho)$,
where $\pi t=\mbox{sgn}(t)\cdot \min\{1,|t|\}$ denotes the truncation of $t\in \mathbb R$ to $[-1,1]$, and $\mbox{sgn}(t)$ is defined as follows
\[
\mbox{sgn} (t):=
\left\{\begin{array}{cc}
1, & \mbox{if} \ t\geq0,\\
-1, & \mbox{if}\ t<0.
\end{array}
\right.
\]
Define the empirical version of $\mathcal E(f)$ as $\mathcal E_D(f):=\frac1m\sum_{i=1}^m\phi(y_if(x_i))$.
Then, we can deduce the following error decomposition easily. Define further $\pi \mathcal H_{\eta,n}:=\{\pi f:f\in\mathcal H_{\eta,n}\}$ and
$f_\phi:=\arg\min_{f\in\pi \mathcal H_{\eta,n}}\mathcal E(f)$. Then, we can derive the following error decomposition.

\begin{lemma}\label{Lemma:Error Decomposition}
Let $f_{D,n}$ be defined by \eqref{new model}.
Then for $f_0\in\mathcal H_{\eta,n}$ with $\|f_0\|_\infty\leq 1$, there holds
\begin{equation}
\label{error decomposition}
   \mathcal E(\pi f_{D,n})-\mathcal E(f_\rho)\leq
   \mathcal E(f_0)-\mathcal E(f_\rho)+\mathcal E(\pi f_{D,n})-\mathcal E(f_\phi).
\end{equation}
\end{lemma}

Our first tool is to construct a $f_0\in\mathcal H_{\eta,n}$ satisfying $\|f_0\|_\infty\leq 1$ with good approximation performance. For this purpose, we recall a construction in the nice paper \cite[Proposition 3]{Tong2016} and get the following lemma.

\begin{lemma}
\label{lemma:approximation-err}
If $\rho$ satisfies   \eqref{Geometric_noise}, then
\begin{equation}\label{approximation-err}
   {\cal E}(B_s(f_{\rho})) - {\cal E}(f_{\rho}) \leq 4 c_1 5^{\alpha} s^{-\alpha},
\end{equation}
where $B_s(f)$ is the   Bernstein polynomial for a function $f$ on the simplex $X$ defined as
\[
B_s(f)(x):= B_{s,d}(f,x) = \sum_{|k|\leq s} f(\frac{k}{s})P_{k,s}(x), \quad x\in X,
\]
where $P_{k,s}(x) = \left(^{s+d}_{\ d}\right)x^k(1-|x|)^{s-|k|}$.
\end{lemma}

%
%\begin{proposition}
%\label{proposition:sampleerror1}
%If $\rho$ satisfies  \eqref{Tsybakov noise}, then for any $0<\delta <1$, with the confidence at least $1-\delta/2$, there holds
%\[
%{\cal S}_D(B_s(f_{\rho})) \leq \frac{8\log(2/\delta)}{3m} + \left( \frac{2c_q \log(2/\delta)}{m}\right)^{\frac{q+1}{q+2}} + \frac{1}{2} D(B_s(f_{\rho})).
%\]
%\end{proposition}
It is obvious that $f_0=D(B_s(f_{\rho}))$ satisfies $\|f_0\|_\infty\leq 1$.
The more challenging part in our analysis  is to bound $\mathcal E(\pi f_{D,n})-\mathcal E(f_\phi)$, for which we adopt the approach in the seminal work \cite{Steinwart2007} and \cite{Mendelson2003-si}.
Define
$$
     \mathcal F:=\{h_f= \phi(y  f)-\phi(yf_\rho):f\in \pi \mathcal H_{\eta,n}\}.
$$
Denote ${\cal N}_2(\epsilon,{\cal F},x_1^m)$
as the $\varepsilon$-covering number of $\mathcal F$ under the   $\ell^2$ norm \cite{Gyorfy2002}[Chap.9], where $x_1^m:=\{x_1,\dots,x_m\}.$ Our first tool is the  error bound for   ${\cal N}_2(\epsilon,{\cal F},x_1^m)$. The main novelty in our analysis, compared with the classical results in \cite{Steinwart2007,Tong2008,Tong2016,Lin2017-GaussSVM} is that we do not impose any restrictions on the RKHS norm of functions in $\mathcal F$. To this end, we should recall the known pseudo-dimension and describes the relation between pseudo-dimension and covering numbers.

If a vector ${\bf t}=(t_1,\dots,t_n)$ belongs to $\mathbb R^n$,
then we denote by $\mbox{sgn} ({\bf t})$ the vector $(\mbox{sgn} (t_1), \dots , \mbox{sgn} (t_n))$.
The VC dimension \cite{Gyorfy2002} of a set $\mathcal V$ over $X$, denoted as $VCdim(\mathcal V)$,
is defined as the maximal natural number $l$ such that there exists a collection $(\xi_1,\dots,\xi_l)$ in  $X$ such that the cardinality
of the sgn-vector  set
\[
S=\{(\mbox{sgn} (v(\xi_1)), \dots, \mbox{sgn} (v(\xi_l))): v\in \mathcal V\}
\]
equals to $2^l$, that is, the set $S$ coincides with the set of all vertexes of unit cube in  $\mathbb{R}^l$.
The quantity
\[
Pdim(\mathcal V):=\max_{g} VCdim(\mathcal V+g),
\]
is called the pseudo-dimension \cite{Maiorov1999} of the set $\mathcal V$ over $X$,
where $g$ runs over the set of all functions defined on $X$ and $\mathcal V+g=\{v+g:v\in \mathcal V\}$.
The following lemma which can be found in  \cite{Mendelson2003-si} establishes some important relations between the pseudo-dimension and covering numbers.

\begin{lemma}\label{Lemma:CAPACITY RELATION}
Let $\mathcal V_R$ be a class of functions which consists of all functions $f\in \mathcal V$ satisfying $|f(x)|\leq R$ for all
$x\in X$. Then
\[
  \sup_{x_1^m\in X^m}\log {\cal N}_2(\epsilon,\mathcal V_R,x_1^m)\leq \bar{c}_1{Pdim}(\mathcal V_R)\log \left(\frac{R}\varepsilon \right),
\]
where $\bar{c}_1$ is an absolute positive constant.
\end{lemma}

Based on Lemma \ref{Lemma:CAPACITY RELATION}, we present a covering number estimate of  $\mathcal F$.
\begin{lemma}\label{Lemma:Covering-Number}
For any $0<p<2$, there is a constant $c_1'$ depending only on $p$ such that
$$
     \sup_{x_1^m\in X^m}\log {\cal N}_2(\epsilon,\mathcal F,x_1^m)\leq c_1'n\log\frac1{\varepsilon^p}.
$$

\end{lemma}

\begin{proof}
Observing that for any $f_1,f_2\in \pi\mathcal H_{\eta,n}$,
\begin{align*}
& |(\phi(y\pi f_1(x))-\phi(y f_\rho(x)))-(\phi(y\pi f_2(x))-\phi(y f_\rho(x)))|\\
&= | \phi(y \pi f_1(x))- \phi(y \pi f_2(x))|
 \leq |\pi f_1(x)-\pi f_2(x)|,
\end{align*}
we obtain
\[
   \sup_{x_1^m\in X^m}{\cal N}_2(\epsilon,{\cal F},x_1^m)\leq \log \sup_{x_1^m\in X^m}{\cal N}_2(\epsilon, \pi\mathcal H_{\eta,n},x_1^m).
\]
It then sufficient to bound the covering number of $\pi\mathcal H_{\eta,n}$.
Noting $\mathcal H_{\eta,n}$ is a linear space with dimension at most $n$,  it follows from \cite{Maiorov1999} that  ${Pdim}(\mathcal H_{\eta,n})\leq n$. Since $\pi\mathcal H_{\eta,n}\subseteq \mathcal H_{\eta,n}$,
By the definition of pseudo-dimension, we then have ${Pdim}(\pi\mathcal H_{\eta,n})\leq n$ \cite[pp. 297]{Maiorov1999}.
Then Lemma \ref{Lemma:CAPACITY RELATION} with $R=1$ implies
\[
   \sup_{x_1^m\in X^m} \log{\cal N}_2(\epsilon,{\cal F},x_1^m) \leq \bar{c}_1 n\log \left(\frac{1}\varepsilon \right).
\]
But $\log\frac1\varepsilon\leq \bar{c}_2\frac1{\varepsilon^p}$ for any $0<p<2$, where $\bar{c}_2$ is a constant depending only on $p$.
The desired estimate holds.
\end{proof}

Our next tool is an error estimate that can
  be easily deduced from \cite[Theorem 5.1]{Steinwart2007} and \cite[Theorem 5.6]{Steinwart2007}.
\begin{lemma}\label{Lemma:concentration-stein}
Let $\delta \in (0,1)$. Suppose that there are constants $c\geq 0,$ $0<\beta\leq 1$ and $B>0$ with
$\mathbf E[h_f^2]\leq c(\mathbf E[h_f])^\beta$ and $\|h_f\|_\infty\leq B$ for all
$h_f\in\mathcal F$. Furthermore, assume that there are constants $a\geq 1$ and $0<p<2$ with
$$
    \sup_{x_1^m\in\mathcal X^m}\log{\cal N}_2(\epsilon,{\cal F},x_1^m)\leq a\varepsilon^{-p}
$$
for all $\varepsilon>0$. Then there exists a constant $c_p>0$ depending only on $p$ such that with confidence $1-\delta$, there holds
\begin{align*}
    &\mathcal E(\pi f_{D,n})-\mathcal E(f_\phi)
    \leq B \left(\frac{a}{m}\right)^\frac{2}{2+p}+\left(\frac{c\log\frac1\delta}m\right)^\frac{1}{2-\beta}\\
     & +
      c_p B^\frac{2p}{4p-2\beta+\beta p}c^\frac{2-p}{4p-2\beta+\beta p}\left(\frac{a}{m}\right)^\frac{2}{4-2\beta+\beta p}+
    \frac{B}{m}\log\frac1\delta.
\end{align*}
\end{lemma}
We then use the above lemmas to bound $\mathcal E(\pi f_{D,n})-\mathcal E(f_\phi)$.  The
 most important stepping stone is  the following relation between variance and expectation that was derived in \cite[Lemma 6.1]{Steinwart2007} under the Tsybakov noise condition (\ref{Tsybakov noise}), that is,
\begin{equation}\label{key-stone}
      \mathbf{E}h_f^2(z)\leq c'_2(\mathbf{E}h_f(z))^{q/(q+1)},
\end{equation}
where $c'_2$ is a  constant depending only on $q$.

With the above tools, we can prove Theorem \ref{Thm:main} as follows.

\begin{proof}{[Proof of Theorem \ref{Thm:main}]} According to (\ref{key-stone}), setting $c=c'_2$, $\beta=\frac{q}{q+1}$, $B=2\|f_\rho\|_\infty$ and $a=c_1'n=c_3's^d$ with $c_3'$ a constant depending only on $p$ and $d$, we get from Lemma \ref{Lemma:concentration-stein} that for any $0<p<2$, with confidence $1-\delta$,
\begin{eqnarray*}
    \mathcal E(\pi f_{D,n})-\mathcal E(f_\phi)
    \leq  c_4' \left(\frac{s^d}{m}\right)^{(q+1)/(q+2+pq/2)}\log\frac1\delta,
\end{eqnarray*}
where $c_4'$ is a constant depending only $p$, $d$ and $\|f_\rho\|_\infty$. Plugging the above estimate and (\ref{approximation-err}) into (\ref{error decomposition}), we obtain for any $0<p<2$, with confidence $1-\delta$, there holds
\begin{align*}
   \mathcal E(\pi f_{D,n})-\mathcal E(f_\rho)
   &\leq 4 c_1 5^{\alpha} s^{-\alpha}\\
   &+c_4' \left(\frac{s^d}{m}\right)^{(q+1)/(q+2+pq/2)} \log\frac1\delta.
\end{align*}
Thus, for $s \sim m^{\frac{q+1}{\alpha(q+2+pq/2) + d(q+1)}}$, we have
$$
   \mathcal E(\pi f_{D,n})-\mathcal E(f_\rho)
   \leq c_5'm^{-\frac{\alpha(q+1)}{\alpha(q+2+qp/2)+d(q+1)}},
$$
where $c_5'$ is a constant depending only on $p$, $d$, $\alpha$ and $\|f_\rho\|_\infty$.
Then Theorem \ref{Thm:main} follows from (\ref{comparison teorem2}).
\end{proof}

\section*{Appendix D. Proof of Lemma \ref{Lemm:solution-hinge-min}}
\label{Appendix:Proof-Hinge-Prox}

\begin{proof}
The identity (\ref{analytic-uk}) can be derived from (\ref{Eq:update-u}) directly. Define
\begin{align*}
&\mathrm{Hinge}_{\gamma}(\xi,\zeta)\\
&=\arg\min_{v\in \mathbb{R}^m} \left\{ \sum_{i=1}^m (1-\xi(i)\cdot v(i))_+ + \frac{\gamma}{2}\|v-\zeta\|_2^2 \right\}
\end{align*}
We can derive (\ref{analytic-vk}) directly.
Let
\begin{align*}
\mathrm{hinge}_{\gamma}(a,b) = \arg\min_{z\in \mathbb{R}} \left\{\max\{0,1-a\cdot z\} + \frac{\gamma}{2}(z-b)^2\right\},
\end{align*}
for some $\gamma>0$ and $a,b\in \mathbb{R}$, then we have
\begin{align*}
&\mathrm{Hinge}_{\gamma}(\xi,\zeta) \\
&= (\mathrm{hinge}_{\gamma}(\xi(1),\zeta(1)), \ldots, \mathrm{hinge}_{\gamma}(\xi(m),\zeta(m)))^T,
\end{align*}
and
\begin{align*}
&\mathrm{hinge}_{\gamma}(a,b) = \\
&\left\{
\begin{array}{cl}
b, &\ \mathrm{if} \ a=0,\\
b+\gamma^{-1}a, &\ \mathrm{if} \ a \neq 0 \ \mathrm{and}\ ab\leq 1-\gamma^{-1}a^2,\\
a^{-1}, &\ \mathrm{if} \ a \neq 0 \ \mathrm{and}\ 1-\gamma^{-1}a^2 <ab<1,\\
b, &\ \mathrm{if} \ a \neq 0 \ \mathrm{and}\ ab\geq 1.\\
\end{array}
\right.
\end{align*}
The only   remainder is to prove (\ref{analytic-for-hinge}).
Given $a,b\in \mathbb{R}$, and $\gamma>0$,
let
\[g(z) := \max\{0,1-a\cdot z\} + \frac{\gamma}{2}(z-b)^2.\]
We consider the minimization problem in the following three different cases: (1) $a>0$, (2) $a=0$ and (3) $a<0$.

\textbf{Case 1. $a>0$:}
In this case,
\begin{align*}
g(z)
= \left\{
\begin{array}{cl}
1-az+\frac{\gamma}{2}(z-b)^2, \ & z<a^{-1},\\
\frac{\gamma}{2}(z-b)^2, \ & z\geq a^{-1}.
\end{array}
\right.
\end{align*}
It is easy to show that the solution of the problem is
\begin{align*}
%\label{Eq:hing-sol-part1}
z^* = \left\{
\begin{array}{cl}
b+\gamma^{-1}a, &\ \mathrm{if} \ a >0 \ \mathrm{and}\ b\leq a^{-1}-\gamma^{-1}a,\\
a^{-1}, &\ \mathrm{if} \ a >0 \ \mathrm{and}\ a^{-1}-\gamma^{-1}a <b<a^{-1},\\
b, &\ \mathrm{if} \ a >0 \ \mathrm{and}\ b\geq a^{-1}.
\end{array}
\right.
\end{align*}

\textbf{Case 2. $a=0$:}
It is obvious that
\begin{align*}
%\label{Eq:hing-sol-part2}
z^* =b.
\end{align*}

\textbf{Case 3. $a<0$:}
Similar to \textbf{Case 1},
\begin{align*}
g(z)
= \left\{
\begin{array}{cl}
1-az+\frac{\gamma}{2}(z-b)^2, \ & z\geq a^{-1},\\
\frac{\gamma}{2}(z-b)^2, \ & {z < a^{-1}}.
\end{array}
\right.
\end{align*}
Similarly, it is easy to show that the solution of the problem is
\begin{align*}
%\label{Eq:hing-sol-part3}
z^* = \left\{
\begin{array}{cl}
b+\gamma^{-1}a, &\ \mathrm{if} \ a <0 \ \mathrm{and}\ b\geq a^{-1}-\gamma^{-1}a,\\
a^{-1}, &\ \mathrm{if} \ a <0 \ \mathrm{and}\ a^{-1}<b<a^{-1}-\gamma^{-1}a ,\\
b, &\ \mathrm{if} \ a <0 \ \mathrm{and}\ b \leq a^{-1}.\\
\end{array}
\right.
\end{align*}
Thus, we finish the proof of this lemma.
\end{proof}

\section*{Appendix E. Proof of Proposition \ref{Propos:dual-form}}
\label{Appendix:Proof-Dual-Form}

\begin{proof}
It is obvious that the problem \eqref{Eq:opt_prob} is equivalent to the following constrained optimization problem
\begin{align*}
& \min_{u \in \mathbb{R}^n,\xi \in \mathbb{R}^m} \ \frac{1}{m}\sum_{i=1}^m \xi_i \\
& \mathrm{s.t.}\ y_i\sum_{j=1}^n A_{ij}u_j \geq 1-\xi_i, \ \xi_i \geq 0, \ i=1,2,\ldots, m.
\end{align*}
The Lagrangian function of the above problem is
\begin{align}
\label{Eq:LagFun}
{\cal L}(u, \xi, {\bf a}, {\bf c})
&= \frac{1}{m}\sum_{i=1}^m \xi_i + \sum_{i=1}^m a_i\left(1-\xi_i - y_i\sum_{j=1}^n A_{ij}u_j\right) \nonumber\\
&- \sum_{i=1}^m c_i\xi_i,
\end{align}
where $a_i, c_i \geq 0$, $i=1,\ldots, m$ are the multipliers.
Let ${\bf a}:= (a_1,a_2,\ldots, a_m)^T$, ${\bf c}:= (c_1, c_2,\ldots, c_m)^T$.
Taking derivatives of the Lagrangian function ${\cal L}$ with $u$ and $\xi$, and letting them equal to zero yields
\begin{align}
& A^T \mathrm{Diag}(y){\bf a} = 0, \label{Eq:derivative-u}\\
& {\bf a} + {\bf c} = \frac{1}{m}{\bf 1}_m. \label{Eq:derivative-xi}
\end{align}
Plugging the above two equations into \eqref{Eq:LagFun}, then the Lagrangian function ${\cal L}$ \eqref{Eq:LagFun} becomes ${\bf 1}_m^T {\bf a}$.
This, together with the equations \eqref{Eq:derivative-u} and \eqref{Eq:derivative-xi} yield the dual form \eqref{Eq:dual-form} presented in Proposition \ref{Propos:dual-form}.
\end{proof}

\section*{Appendix F. Sensitivity Study of Parameters}
\label{Appendix:Sensitivity-para}

In Algorithm \ref{alg1}, besides the degree $s$ of polynomial kernel, there are three parameters, i.e., the center generation mechanism, the proximal parameter $\alpha$ and penalty parameter $\beta$ used in the implementation of ADMM algorithm. In this section, we conducted a series of studies on their sensitivity over UCI data sets.

\begin{figure}[!t]
\begin{minipage}[b]{0.49\linewidth}
\centering
\includegraphics*[scale=0.33]{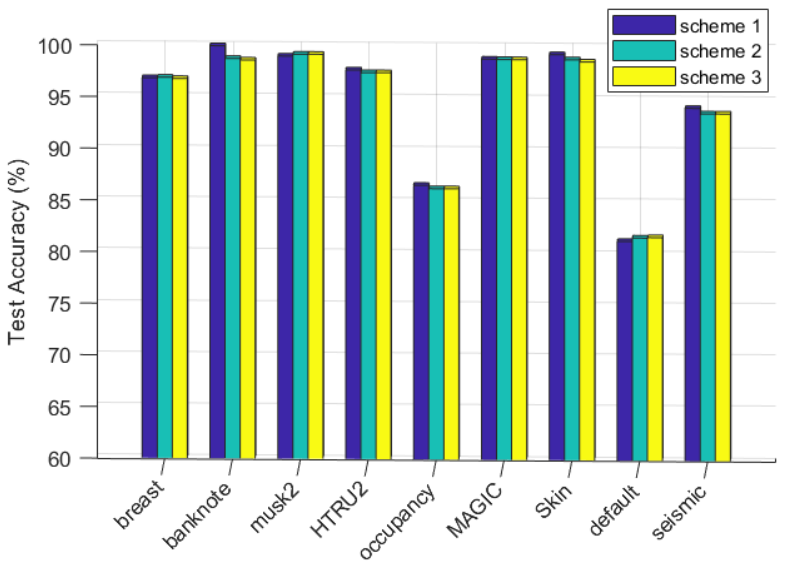}
%  \vspace{-.5cm}
\centerline{{\small (a) Test accuracy}}
\end{minipage}
\hfill
\begin{minipage}[b]{0.49\linewidth}
\centering
\includegraphics*[scale=0.33]{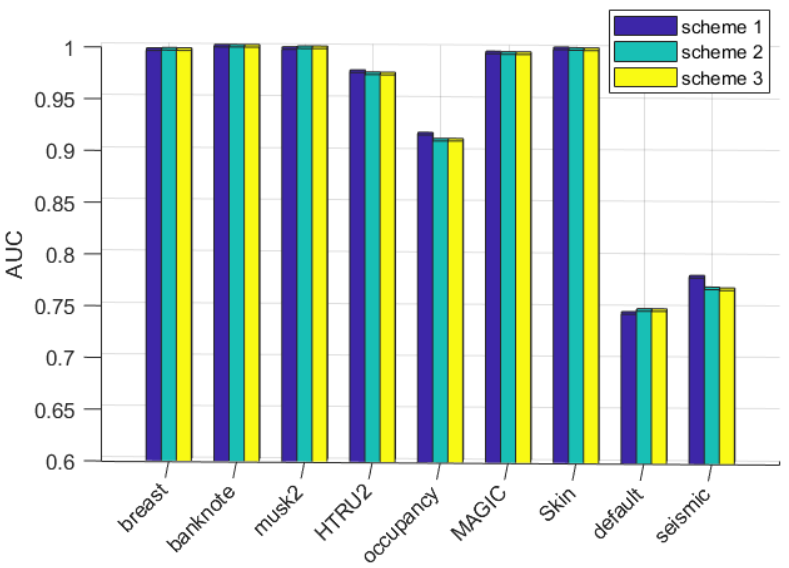}
%  \vspace{-.5cm}
\centerline{{\small (b) AUC}}
\end{minipage}
\hfill
\caption{Sensitivity of center generation mechanisms for FPC over UCI data sets.
 }
\label{fig:sensitivity-center}
\end{figure}

{\bf 1) On center generation mechanism.}
In this part, we considered three different center generation mechanisms as presented in {\bf Simulation 5} in Section V.B. For all UCI data sets, the parameters $\alpha$ and $\beta$ were fixed to be 1 in these experiments, and the degree of polynomial kernel $s$ was empirically selected from the range $(1, s_{\max})$, where
$s_{\max}:= \min\left\{[m^{1/d}], 10\right\}$.
%$s_{\max}:= \min\left\{\left\lceil (\frac{m}{\log m})^{1/d}\right\rceil, 10\right\}$.
For a pre-determined $s$, the number of centers $n$ was set as $n= \min\left\{\left(^{s+d}_{\ s}\right),m\right\}$. The experiment results are shown in Fig. \ref{fig:sensitivity-center}. From Fig. \ref{fig:sensitivity-center}, the performance of all three mechanisms are almost the same in terms of both test accuracy and AUC. This is also verified by our developed theory, i.e., Proposition \ref{Proposition:INDEPENDENCE}. In terms of test accuracy, the performance of FPC equipped with scheme 1, i.e., generating the center points via a uniformly random way is slightly better than the other two schemes in average. Thus, we suggest using scheme 1 for FPC in experiment.

\begin{figure}[!t]
\begin{minipage}[b]{0.49\linewidth}
\centering
\includegraphics*[scale=0.33]{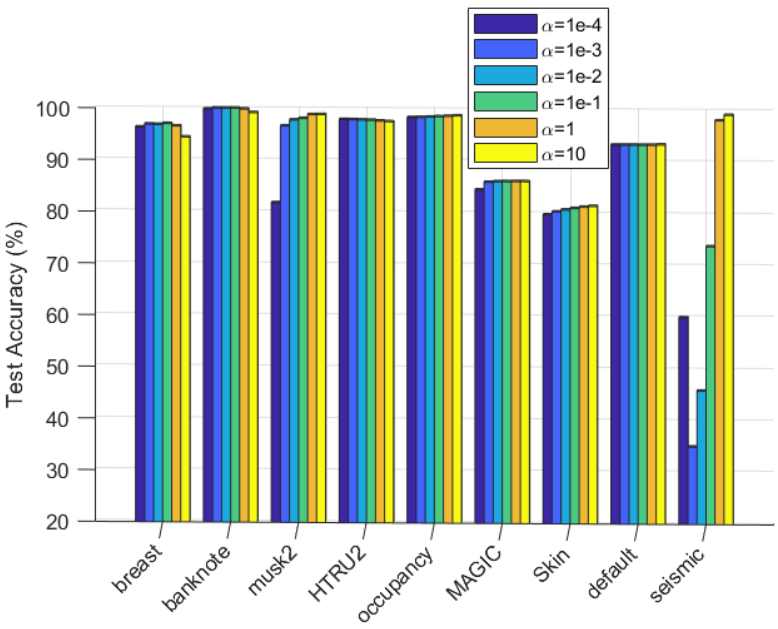}
%  \vspace{-.5cm}
\centerline{{\small (a) Test accuracy}}
\end{minipage}
\hfill
\begin{minipage}[b]{0.49\linewidth}
\centering
\includegraphics*[scale=0.33]{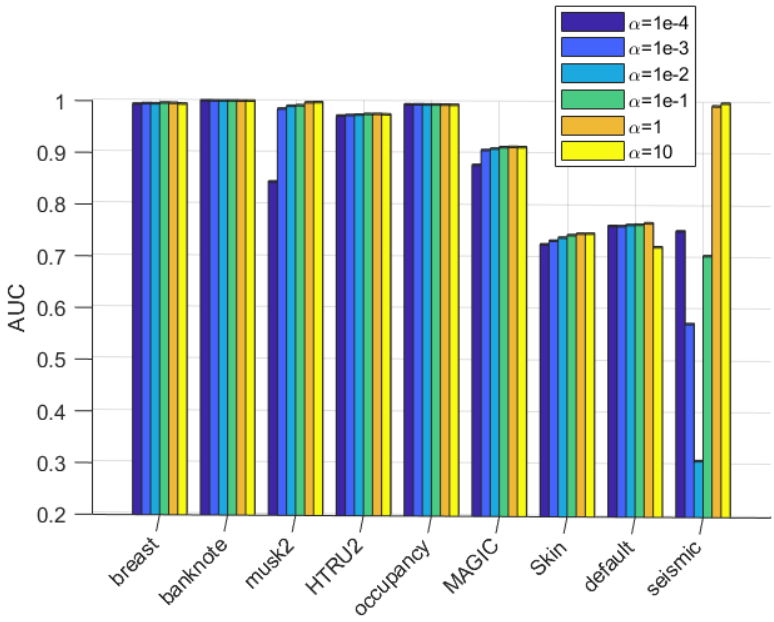}
%  \vspace{-.5cm}
\centerline{{\small (b) AUC}}
\end{minipage}
\hfill
\caption{Sensitivity of proximal parameter $\alpha$ for FPC over UCI data sets.
 }
\label{fig:sensitivity-proximal}
\end{figure}

\begin{figure}[!t]
\begin{minipage}[b]{0.49\linewidth}
\centering
\includegraphics*[scale=0.33]{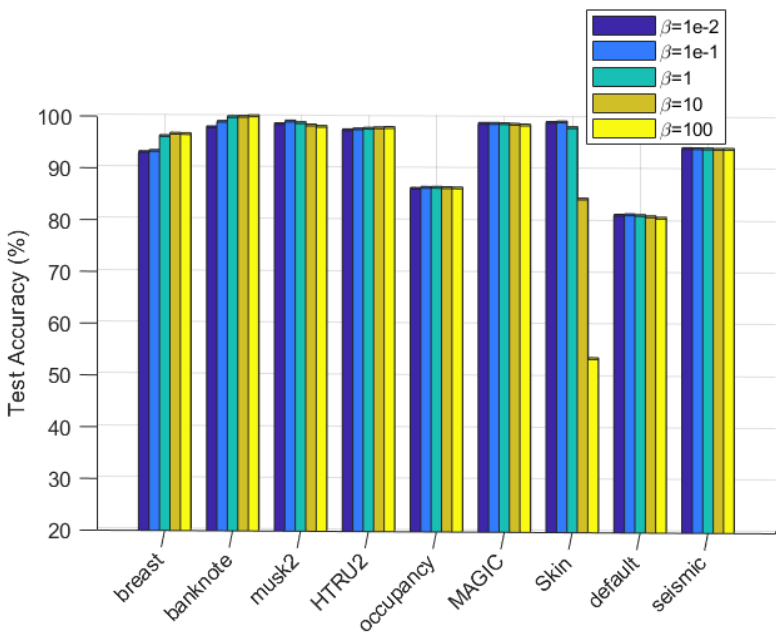}
%  \vspace{-.5cm}
\centerline{{\small (a) Test accuracy}}
\end{minipage}
\hfill
\begin{minipage}[b]{0.49\linewidth}
\centering
\includegraphics*[scale=0.33]{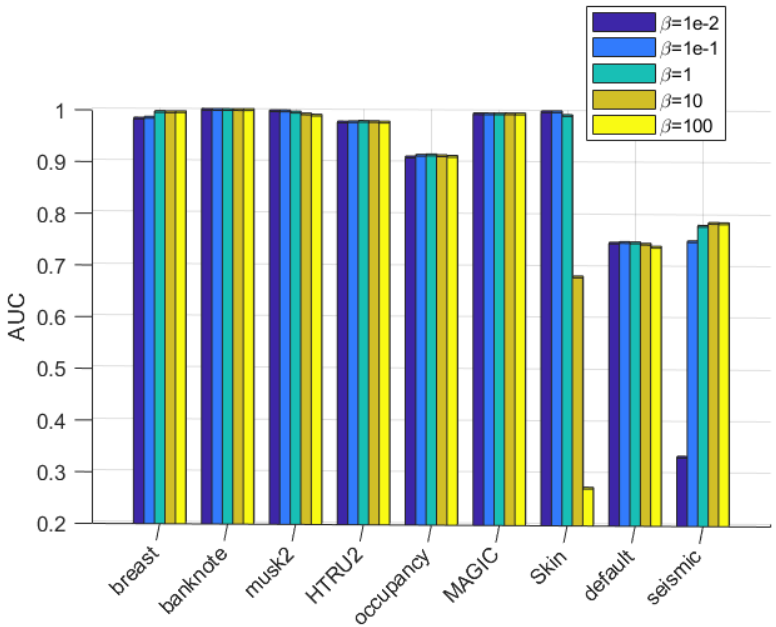}
%  \vspace{-.5cm}
\centerline{{\small (b) AUC}}
\end{minipage}
\hfill
\caption{Sensitivity of penalty parameter $\beta$ for FPC over UCI data sets.
 }
\label{fig:sensitivity-penalty}
\end{figure}

{\bf 2) On proximal parameter $\alpha$.} In these experiments, we considered the sensitivity of proximal parameter $\alpha$ for FPC over UCI data sets. The experimental settings were similar to Appendix F.A, except that only scheme 1 was used in FPC and six different $\alpha$'s are considered, i.e., $\{$1e-4, 1e-3, 1e-2, 1e-1, 1, 10$\}$. The experiment results are shown in Fig. \ref{fig:sensitivity-proximal}. From Fig. \ref{fig:sensitivity-proximal}, we can observe that FPC is generally stable to the proximal parameter for most of the UCI data sets, and particularly, FPC with $\alpha=1$ achieves the best performance in average over these UCI data sets. Thus, in the real world data experiments, we suggest using $\alpha =1$ as the default value for FPC in experiments.

{\bf 3) On penalty parameter $\beta$.}
The experiment settings in this part were similar to Appendix F.B, except that $\alpha$ was fixed to be 1 and $\beta$ varied from $\{$1e-2, 1e-1, 1, 10, 100$\}$.
The experiment results on the sensitivity of the penalty parameter $\beta$ are shown in Fig. \ref{fig:sensitivity-penalty}.
From Fig. \ref{fig:sensitivity-penalty}, we can observe that $\beta =1$ is almost the best for FPC over all nine UCI data sets. Therefore, we suggest using $\beta=1$ as the default value for FPC in experiments.

%\bibliographystyle{IEEEtran}
%% \bibliography{reference}
%\bibliography{myreferences}

%% biography section
%%
%% If you have an EPS/PDF photo (graphicx package needed) extra braces are
%% needed around the contents of the optional argument to biography to prevent
%% the LaTeX parser from getting confused when it sees the complicated
%% \includegraphics command within an optional argument. (You could create
%% your own custom macro containing the \includegraphics command to make things
%% simpler here.)
%%\begin{IEEEbiography}[{\includegraphics[width=1in,height=1.25in,clip,keepaspectratio]{mshell}}]{Michael Shell}
%% or if you just want to reserve a space for a photo:
%\end{IEEEbiography}
%% insert where needed to balance the two columns on the last page with
%%\newpage


%\vfill


\end{document}